\documentclass{article}

\usepackage{genai4health_2026}
\usepackage[utf8]{inputenc} 
\usepackage[T1]{fontenc}    
\usepackage{hyperref}       
\usepackage{url}            
\usepackage{booktabs}       
\usepackage{amsfonts}       
\usepackage{nicefrac}       
\usepackage{microtype}      
\usepackage{xcolor}         

\usepackage{graphicx}
\usepackage{hyperref}
\usepackage{colortbl}
\usepackage{siunitx}
\usepackage{makecell}
\usepackage{fvextra}
\usepackage{longtable}
\usepackage{xcolor}
\usepackage{xltabular}
\AtBeginDocument{\setlength{\LTcapwidth}{\textwidth}}
\usepackage{etoolbox}
\makeatletter
\patchcmd{\LT@makecaption}{\reset@font}{\reset@font\normalsize}{}{\@latex@warning{longtable caption font patch failed}}
\makeatother
\usepackage[]{todonotes}  
\setuptodonotes{inline}
\usepackage{xurl}
\usepackage{booktabs}
\usepackage{cleveref}
\usepackage{enumitem}
\usepackage{multirow}
\usepackage{graphicx}
\graphicspath{{images/}}
\usepackage{pgffor}
\usepackage{array}
\usepackage{xspace}
\usepackage{mdframed}
\usepackage{wrapfig}  
\usepackage{capt-of}  
\usepackage{ragged2e}

\newcommand{\pipeline}{\texttt{Retrieval-based factuality evaluation}\xspace}

\title{Where Does Retrieval-Based Open-Ended Evaluation Fail? Automatic Taxonomy Induction from Long-Form Medical Answer Factuality Verification}

\author{Heyuan Huang\thanks{Equal contribution}
\quad Jirui Dai$^{*}$ \quad Alexandra DeLucia \quad Sonal Joshi \\ {\bf Mahsa Yarmohammadi \quad Jie Gao \quad Bernal Jiménez Gutiérrez \quad Mark Dredze} \\
        Center for Language and Speech Processing \\ Johns Hopkins University \\ Baltimore, MD 21218, USA\\
        \texttt{\{hhuan134,  mdredze\}@jhu.edu}}

\begin{document}

\maketitle

\begin{abstract}
Retrieval-based factuality evaluation, where LLM-generated claims are verified against evidence from authoritative medical corpora, has become the dominant paradigm for scalable hallucination detection in high-stakes clinical settings.
Despite the urgency of reliable and transparent medical fact verification, most systems measure performance with aggregate metrics like F1, which obscure where and why failures occur. Existing RAG diagnostics require gold answers or annotated gold evidence, neither of which exists in this regime.
We introduce two comprehensive taxonomies, grounded in a case study on the open-ended MedExpert dataset and 3 closed-ended datasets, decomposing failures into retrieval-stage errors along five quality dimensions, and verifier-reasoning errors into six consecutive steps.
We adapt an automatic pattern induction pipeline using LLM-as-Judge to label evidence quality and classify verifier reasoning errors at scale, and then stress-test our findings across 4 retrieval methods and 6 frontier verifier models.
Our analysis reveals that scaling model size, adding reasoning effort, expanding to authoritative web sources, and applying medical fine-tuning do not resolve these failure modes, demonstrating that they represent fundamental limitations of the retrieve-then-verify paradigm in open-ended medical settings rather than artifacts of outdated systems.\footnote{We release our code and data at \url{https://github.com/Heyuan9/Open-Ended_RAG_Evaluation} for the full reproducibility of our results.}
\end{abstract}

\section{Introduction}
The rapid adoption of Large Language Models (LLMs) in patient-facing Question Answering (QA), report generation, and summarization has necessitated rigorous frameworks to evaluate the factuality of
  generated text. To detect plausible but unfactual information, \citet{min-etal-2023-factscore} proposed a decompose-then-verify paradigm, FActScore, to granularly check each atomic claim against a knowledge
   source. Building on this paradigm, MedScore \citep{medscore} decomposes complex medical answers into independent, condition-aware claims and verifies them against passages retrieved from the MedRAG corpus
  \citep{xiong-etal-2024-benchmarking}. This widely adopted paradigm is built on the premise that retrieved evidence from authoritative corpora is sufficiently factual and relevant.

  However, our experiments reveal that even SoTA pipelines fail severely in this open-ended medical setting: on MedExpert \citep{yarmohammadi2025medexpert}, an open-ended medical QA dataset annotated by
  clinicians, pairing the strongest retrievers with frontier verifiers recovers only a small fraction of clinician-flagged factual errors --- an order-of-magnitude collapse from the same pipeline's
  performance on closed-ended biomedical benchmarks. This collapse persists across model scaling, increased reasoning effort, medical fine-tuning, and corpus expansion to authoritative web sources, pointing
  to a \emph{structural} limitation of the retrieve-then-verify paradigm in the open-ended regime rather than an artifact of outdated systems. End-to-end metrics like F1, however, cannot say \emph{where} or
  \emph{why} the pipeline breaks.

  To open this black box, we present a \textbf{comprehensive model-agnostic error taxonomy} for retrieval-based open-ended factuality evaluation, decomposing failures into two stages:

  \begin{itemize}[nosep]
      \item \textbf{Retrieval-stage errors}: irrelevant evidence, mismatch clinical scope,  trustworthiness, evidence polarity, and chunking issues.
      \item \textbf{Verification-stage errors}: incorrect evidence selection, failure in evidence comprehension, insufficient reasoning, overconfidence, inconsistency, and hallucinations.
  \end{itemize}
\begin{wrapfigure}{R}{0.55\linewidth}
    \centering
    \vspace{-1.5\intextsep}
    \includegraphics[width=\linewidth]{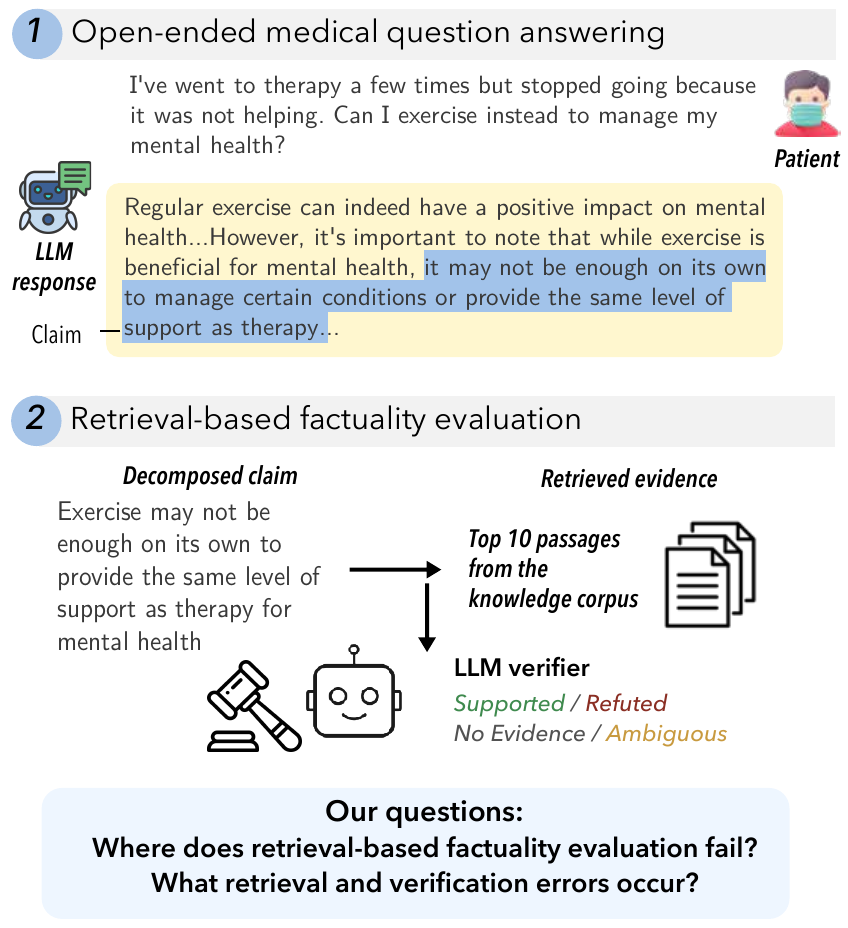}
    \caption{Retrieval-based factuality evaluation of open-ended medical answers: an LLM verifier checks claims against retrieved evidence, motivating our evaluation of where this process fails for Stage-wise reliability.}
    \label{fig:motivation}
    \vspace{-55pt}
\end{wrapfigure}
  The taxonomy is induced via a human-in-the-loop LLM-as-judge pipeline that requires no gold answers or annotated gold evidence, a regime where existing closed-ended RAG diagnostics \citep{ru2024ragchecker, es-etal-2024-ragas, leung-etal-2026-classifying,
  sivakumar2026ragx} do not apply, and that scales diagnosis far beyond manual inspection.
  
{\RaggedRight\hyphenpenalty=10000
We summarize our contributions as follows.
\begin{enumerate}[leftmargin=*]
      \item A multi-dimensional error taxonomy for retrieval-based open-ended factuality evaluation, automatically inducible without gold annotation.
      \item A large-scale empirical analysis across 6 verifier LLMs, 4 retrievers, and 3 knowledge corpora, showing that scaling, reasoning effort, medical fine-tuning, and corpus expansion all fail to
  resolve the dominant failure patterns.
      \item Actionable recommendations for researchers to improve open-ended factuality systems: 
      A multi-dimensional evaluation protocol, including task-oriented assessment of retrieved evidence paired with intermediate-step assessment of verifier behavior, reflects system capability more faithfully and surfaces failure modes more directly than aggregated end-to-end metrics. 
\end{enumerate}}

\begin{figure*}[!htbp]
    \centering
    \includegraphics[width=\linewidth]{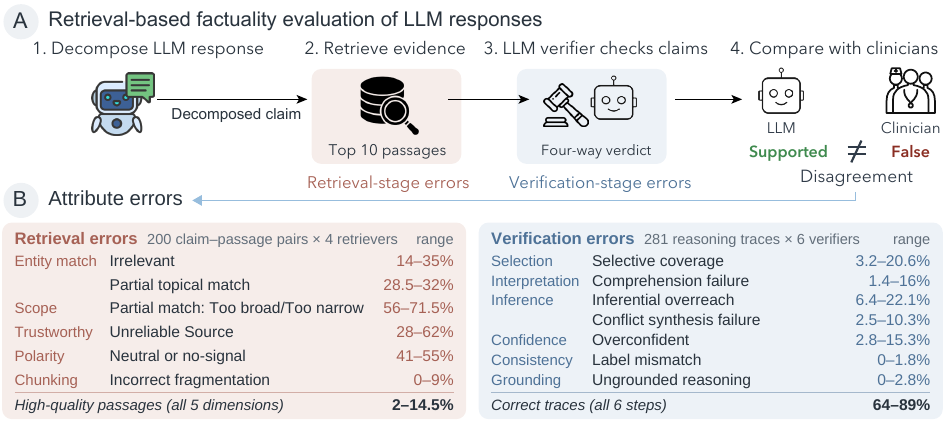}
    \caption{Diagnosing error modes in a retrieve-then-verify pipeline: (A) A pipeline decomposes a response sentence into atomic facts, retrieves top-10 medical passages, asks an LLM verifier for a four-way label (Supported / Refuted / No Evidence / Ambiguous), and compares it against a human verdict (True/False). (B) Our contribution is to automatically induce a taxonomy of retrieval error modes (where the evidence itself is the bottleneck) and verification error modes (where the LLM reasons incorrectly over the evidence). Full error modes are in \Cref{tab:codebook_subcode_by_retriever} and \Cref{tab:codebook_subcode_by_verifier}. Ranges span each pattern's lowest to highest rate across the 4 retrievers (200 passages each) or 6 verifiers (281 reasoning traces each) on MedExpert-gold.}
    \label{fig:error-overview}

\end{figure*}
\section{Related Work}

\paragraph{Open-ended factuality evaluation.}
Evaluating the factuality of diverse text generated by LLMs remains a persistent challenge, especially when it is open-ended. Traditional n-gram-based metrics require costly human annotation, while recent annotation-free approaches---FActScore \citep{min-etal-2023-factscore}, and MedScore \citep{medscore}---retrieve text snippets from external sources (e.g., Wikipedia, PubMed) as reference for automated verification. These systems use LLM verifiers to generate a verdict for each atomic claim based on the retrieved evidence, which is not 100\% reliable due to the evidence quality and capabilities of LLMs. However, these decompose-then-verify pipelines only output end-to-end aggregated factuality scores and verifiers' reasoning in plain text, making it difficult to diagnose \emph{where} the pipeline fails.

\paragraph{Retrieval evaluation in closed- and open-ended fact verification}
  Retrieval quality evaluation for fact verification has matured under the
  \emph{closed-ended} paradigm and remains largely absent in the open-ended
  domain. Closed-ended fact verification benchmarks---FEVER \citep{thorne-etal-2018-fever},
  SciFact \citep{wadden-etal-2020-fact},
  HealthVer \citep{sarrouti-etal-2021-evidence-based},
  CovidFact \citep{saakyan-etal-2021-covid},
  PubHealth \citep{kotonya-toni-2020-explainable}---curate each claim, identify the specific passages that constitute its valid evidence, and then annotate its gold verdict (Supported / Refuted / Not Enough Info). With gold evidence
  annotated per claim, the retriever is evaluated directly by the evidence
  precision@$k$ and recall@$k$, and verifier errors can be cleanly separated
  from retrieval misses; on top of this protocol, a mature line of
  diagnostic frameworks measure the quality of the retrieve-then-verify pipelines by stage
  \citep{ru2024ragchecker, es-etal-2024-ragas, leung-etal-2026-classifying,
  sivakumar2026ragx}. 
  
  Open-ended fact verification has none of this scaffolding. In open-ended \pipeline, the claim is \emph{decomposed} from an LLM's long-form generation rather than curated; there is no annotated verdict to compare against, and no annotated gold passages---LLM-generated atomic claims are too diverse to enumerate evidence for, and a claim may itself be a parametric fabrication with no factual support in \emph{any} corpus, making evidence recall@$k$ undefined in principle, rather than merely expensive to annotate. Existing open-ended factuality pipelines, therefore, treat retrieval as a black box
  subroutine and score only the final claim verdict \citep{min-etal-2023-factscore, medscore}; no standardized metric, framework, or benchmark evaluates the retriever itself. 
  
  Two recent medical-RAG diagnostics remain closed-ended. MedRAGChecker \citep{ji2026medragchecker}, a medical-domain adaptation of RAGChecker \citep{ru2024ragchecker}, uses gold answers' claims as ground truth to calculate claims recalled by retrieved evidence, and can only be used when there exist gold answers. \citet{kim2025rethinkingRAGinmedicine} evaluates retrieval via post-hoc relevance annotations on retrieved passages, but anchors them to physician-written answers' must-have statements. We provide the first systematic error analysis of the retrieval \emph{stage} when no reference answer exists at all.

Our work focuses on \emph{error taxonomy construction}---systematically categorizing \emph{why} each stage fails---rather than building a better fact verification system, and on using \textit{automatic} LLM-based error pattern extraction to scale the analysis beyond manual inspection.

\section{Experimental Setup}
\subsection{Open-ended Datasets}

\textbf{MedExpert} \citep{yarmohammadi2025medexpert} contains 540 open-ended medical QA pairs that are annotated for factuality and corresponding severity of the unfactual information by clinicians within their Mental Health (MH) and Prenatal Care (PC) specialties. Clinicians highlight unfactual text spans, label them as "False", and write reasons. For each sentence with an unfactual span, we consider it a Positive case (i.e., contains factual errors). We use MedScore \citep{medscore} to decompose each response into context-aware claims and verify each claim against retrieved evidence. For manual evaluation, we sampled a gold subset, MedExpert-gold, whose factuality annotation is double-checked by five practicing clinicians with relevant MH and PC clinical licensure. Data representativeness is discussed in \Cref{paragraph:representativeness}. We use MedExpert and MedExpert-gold as the main datasets for this case study and show that our findings are generalizable to the simulated "semi open-ended" datasets below. 

\subsection{Closed-Ended Datasets}

Due to the limited number of publicly available open-ended medical factuality evaluation datasets, we transform 4 closed-ended medical fact check datasets into the open-ended setting by changing their own curated small evidence corpus to our 30.3M passage corpus, MEDIC, and convert the 4 datasets into a
unified \{\texttt{id}, \texttt{claim}\} format.

\textbf{SciFact} \citep{wadden-etal-2020-fact} contains 1,409 expert-written scientific claims derived from biomedical research articles, with evidence abstracts and annotated verdict labels based on the evidence. \textbf{SciFact-Open} \citep{wadden-etal-2022-scifact} extends this setting toward semi-open-domain by using a 500K biomedical scientific abstract corpus as the knowledge base for 279 claims' verification. Both datasets' claims are labelled as Support, Refute, and No Evidence.

\textbf{HealthVer} \citep{sarrouti-etal-2021-evidence-based} contains 1,855 real-world health-related claims collected from search engine results for health questions. Each claim is paired with evidence from scientific articles and annotated with Supports, Refutes, and Neutral verdicts.

\textbf{CovidFact} \citep{saakyan-etal-2021-covid} contains 4,086 COVID-19 claims and filters each claim's top 5 Google Search results into gold evidence and annotates Supported/Refuted verdicts.

Data preprocessing and alignment details are in \Cref{sec:data preprocessing and alignment}, and preprocessed data statistics are in \Cref{tab:data_statistics}.

\subsection{Evaluation Setup}
\label{sec:configurations}

\citet{medscore} finds that binary verification usually assigns "False" to a claim for variable reasons, while it only assigns "True" when there is exact supporting evidence. We extend MedScore's binary verdict to enable finer-grained error analysis. For each medical claim in the 4 datasets, we verify it against retrieved evidence, using a 4-way verdict: \textit{Supported}, \textit{Refuted}, \textit{No Evidence}, and \textit{Ambiguous}. When the retrieved passages provide conflicting evidence, the verifier should assign the label \textit{Ambiguous} to the claim. Such disagreement is common in the medical domain, where different studies, guidelines, or clinical sources may offer diverging conclusions on the same topic.

\subsection{Methods}

\paragraph{Retrieval methods.} We compare 4 popular and mostly used retrievers: 
\begin{itemize}[nosep]
\item \textbf{MedCPT}\citep{jin2023medcpt}, which is the most commonly used medically fine-tuned retriever with 384,891 monthly downloads in April, 2026.\footnote{\url{https://huggingface.co/ncbi/MedCPT-Query-Encoder}}

\item \textbf{A hybrid retriever}, RRF-2, combining BM25 (lexical retriever) and MedCPT (biomedical domain semantic retriever), using Reciprocal Rank Fusion.

\item \textbf{A hybrid retriever}, RRF-4, combining BM25, MedCPT, SPECTER (scientific domain semantic retriever)\citep{cohan-etal-2020-specter}, and Contriever (general domain semantic retriever)\citep{izacard2022unsuperviseddenseinformationretrieval}, using the same Reciprocal Rank Fusion.
\item \textbf{Qwen3-Embedding-8B}\citep{qwen3embedding}, which is the best general domain embedding model in 2025 on the MTEB leaderboard.
\end{itemize}

Each retriever uses the decomposed claim as the query to retrieve the top 10 relevant passages as the verification evidence.

\paragraph{Knowledge corpora.} We compare 3 knowledge sources:
\begin{itemize}[nosep]
\item \textbf{MEDIC corpus}, including 30.3M passages collected by \citet{xiong-etal-2024-benchmarking} from PubMed, StatPearls, and Medical Textbook.\footnote{We excluded its Wikipedia (general knowledge) part to avoid medical-related snippets from lines/dialogues in medical-themed TV dramas and movies.} 

\item \textbf{Google}\footnote{\href{https://serper.dev/}{Serper API} is used for Google search engine} \textbf{general search} as a special case study to see if authoritative domain filtering (e.g.,.edu, .gov, .org) can introduce up-to-date, reliable evidence, compared with static corpus retrieval.

\item \textbf{Google Scholar search} that limits the evidence to articles indexed by the Google Scholar engine, whose scope is broader than the static MEDIC corpus but smaller than Google general search.
\end{itemize}

\paragraph{Verifier LLMs.} We compare 6 verifier LLMs:
\begin{itemize}[nosep]
\item \textbf{GPT-5.4}, one of the SOTA closed-source general domain LLMs \citep{singh2026openaigpt5card}.

\item \textbf{MedGemma 27B}, a medically fine-tuned open-source LLM \citep{sellergren2025medgemma}.

\item \textbf{Gemma 3 27B}, a general domain open-source LLM with the same architecture and pre-training, but different fine-tuning from MedGemma \citep{gemma_2025}.

\item \textbf{Qwen3.6-27B}, one of the SOTA open-source general domain LLMs that achieves comparable benchmark scores with Claude 4.5 Opus \citep{qwen3.6-27b}.\footnote{https://huggingface.co/Qwen/Qwen3.6-27B}

\item \textbf{Mistral Small 3} (24B-Instruct-2501), the original MedScore verifier, as our baseline \citep{mistral_ai_mistral_2025}.

\item \textbf{Mistral Small 4}(119B-2603), a newly released open-source LLM from the Mistral family whose benchmark scores are higher than Mistral Small 3 \citep{mistralai2026mistralsmall4}.\footnote{https://huggingface.co/mistralai/Mistral-Small-4-119B-2603}
\end{itemize}

For each combination of retrieval method, knowledge corpus, and verifier LLM, we run the full verification pipeline and apply automatic error analysis to the results.

\subsection{Automatic Pattern Extraction}
\label{sec:error_extraction}

We aim to evaluate the performance of the \pipeline in comparison to human evaluators (i.e., clinicians), particularly its error patterns. Our goal is to identify as many errors as possible, particularly errors that \pipeline overlooked or misclassified (i.e., false negatives and false positives), and potentially surfacing additional errors that clinicians themselves did not detect. Manual inspection of false positive and false negative cases does not scale. We develop an automatic pipeline using LLM-as-judge. Similar to prior work that employs an LLM as a qualitative judge~\cite{chirkova2025llmasaqualitativejudgeautomatingerroranalysis}, we first developed a preliminary error codebook based on our interactions with clinicians, then used this as a seed codebook to elicit additional error patterns not previously identified. Our seed error codebook is below:

\paragraph{Retrieval-stage seed codebook.} For each claim--passage pair, an LLM judge labels the retrieved evidence along multiple dimensions beyond the traditional relevancy score (e.g., cosine similarity):
\begin{itemize}[nosep]
    \item \textbf{Entity Match}: Direct match; Partial topical match; Irrelevant.
    \item \textbf{Scope}: Whether the evidence addresses the specific clinical subgroup (e.g., pregnant women, pediatric patients) or only provides general population descriptions.
    \item \textbf{Trustworthy}: Whether the content is based on a trustworthy primary trial with a sufficient number of subjects.
    \item \textbf{Evidence Polarity}: Whether this evidence gives direct affirmation or contradiction to the claim, or it is just a topic mention without a proposition.
\end{itemize}

\paragraph{Verifier-stage seed codebook.} Each verifier's reasoning trace is decomposed into 6 consecutive steps and example errors:
\begin{itemize}[nosep]
    \item \textbf{Interpretation}: The verifier misunderstood the claim or the evidence.
    \item \textbf{Evidence selection}: The verifier missed relevant evidence, or focused more on low-quality evidence
    \item \textbf{Inference}: The verifier could not integrate information across multiple passages to reach a correct verdict. The verifier lacked domain knowledge to reason as a clinician would (e.g., understanding drug interactions, contraindications).
    \item \textbf{Calibration}: the verifier is overconfident or over-hedging on the evidence strength
    \item \textbf{Consistency}: the verifier's reasoning does not align with its final verdict
    \item \textbf{Grounding}: the verifier used information that is not provided in the evidence
\end{itemize}

We manually synthesized the saturated final codebooks and validated LLM judge labels against a human-annotated subset and report inter-annotator agreement, with taxonomy construction details in \Cref{sec:taxonomy construction details}.

\begin{table}[t]
\caption{Precision/Recall/F1*100 (\%) under different binary label-mapping strategies across open-ended and semi-open-ended fact-checking datasets, using SOTA retriever, Qwen3 in MEDIC corpus, and SOTA verifier, GPT-5.4. Refuted+NE+A means we count Refuted, No Evidence, and Ambiguous labels as Has Error claims. The highest F1 and Recall are bolded for each dataset.}
\label{tab:all-datasets-results}
\centering
\sisetup{
  table-format=2.1,
  table-number-alignment=center,
  detect-weight=true,
  detect-family=true
}
\resizebox{\textwidth}{!}{%
\begin{tabular}{l *{18}{c}}
\toprule
& \multicolumn{6}{c}{\textbf{Open-Ended}} & \multicolumn{12}{c}{\textbf{Semi-Open-Ended}} \\
\cmidrule(lr){2-7} \cmidrule(lr){8-19}
& \multicolumn{3}{c}{MedExpert} & \multicolumn{3}{c}{MedExpert-gold}
& \multicolumn{3}{c}{CovidFact} & \multicolumn{3}{c}{HealthVer}
& \multicolumn{3}{c}{SciFact-Open} & \multicolumn{3}{c}{SciFact} \\
\cmidrule(lr){2-4} \cmidrule(lr){5-7}
\cmidrule(lr){8-10} \cmidrule(lr){11-13}
\cmidrule(lr){14-16} \cmidrule(lr){17-19}
\textbf{Label Mapping}
& {P} & {R} & {F1} & {P} & {R} & {F1}
& {P} & {R} & {F1} & {P} & {R} & {F1}
& {P} & {R} & {F1} & {P} & {R} & {F1} \\
\midrule
Refuted 
& 5.6 & 17.0 & \textbf{8.4} & 56.4 & 25.6 & 35.2
& 94.8 & 33.7 & 49.7 & 40.4 & 37.5 & 38.9
& 67.3 & 78.7 & \textbf{72.5} & 57.3 & 81.4 & \textbf{67.2} \\
Refuted\,+\,NE 
& 3.2 & 33.9 & 5.9 & 46.8 & 41.9 & 44.2
& 82.9 & 67.8 & 74.6 & 33.0 & 47.8 & 39.0
& 51.3 & 86.5 & 64.4 & 38.6 & 86.9 & 53.4 \\
Refuted\,+\,NE\,+\,A 
& 3.1 & \textbf{45.1} & 5.7 & 47.1 & \textbf{55.8} & \textbf{51.1}
& 80.3 & \textbf{76.3} & \textbf{78.2} & 27.0 & \textbf{74.6} & \textbf{39.6}
& 47.3 & \textbf{98.9} & 64.0 & 35.5 & \textbf{97.0} & 52.0 \\
\bottomrule
\end{tabular}%
}
\end{table}

\section{Open-Ended vs. Closed-Ended Fact Verification Results}
\label{sec:open-endedness}

Our main results in \Cref{tab:all-datasets-results} demonstrate that open-ended retrieval-based fact verification remains fundamentally challenging for the retrieve-then-verify pipeline, given the consistently low Recall and F1 of that even the strongest available models obtain in the true open-ended MedExpert dataset.


This challenge is driven by three fundamental properties of open-ended fact-checking task: (i) the \textbf{comprehensiveness} of the evidence base, (ii) the \textbf{identifiability} of gold evidence within it, and (iii) the \textbf{alignment} between retrieved evidence and the claim. The rising end-to-end F1 we observe across datasets, from $0.08$ on MedExpert to $0.7$ to $0.8$ on CovidFact and SciFact-style benchmarks, shows that the real open-ended tasks have all these 3 difficult properties, making it more challenging than the semi-open-ended task, SciFact-style benchmarks, which almost don't have these 3 difficulties.

\paragraph{Comprehensiveness of the evidence base.}
The most basic requirement is that the gold evidence supporting (or refuting) a claim actually exists in the corpus available to the system. We find that approximately 90\% to 81\% of the gold passages in SciFact and SciFact-Open are contained in MEDIC, and end-to-end F1 on these datasets reaches 0.7 once the relevant evidence is retrieved. In contrast, COVID-related datasets draw their gold passages from web sources --- opinion pieces, news articles, and lay-audience summaries --- that fall outside the scientific literature indexed by MEDIC; F1 on these datasets drops to 0.4 to 0.5. In MedExpert, which is open-ended by construction and offers no guarantee of evidence existence or coverage, F1 collapses to $0.06$. This property can not be solved by retriever or verifier improvement, and can not be solved by simply expanding the corpus, as shown in \Cref{sec:google search discussion}. More details about the overlap between each dataset's gold passages and the MEDIC corpus can be found in \Cref{sec:closed-ended evidence overlap}.

\paragraph{Identifiability of evidence.}
Even when gold evidence is present, it must be locatable. In closed-ended datasets, the corpus was constructed \emph{around} the claims, so the search space is small and curated and the right passages are comparatively easy to surface. In an open-ended setting operating over millions of passages, the same gold evidence is buried among far more distractors, and retrieval quality becomes the binding constraint on end-to-end performance. Unlike comprehensiveness, this is a scaling-friendly problem: as retrievers improve, identifiability gaps shrink, and the corresponding portion of the end-to-end error rate decreases.

\paragraph{Alignment between evidence and claim.}
In the medical domain, claims and evidence are frequently misaligned even when they are topically matched. A claim stating an absolute medication dose (``adults may take 400 mg of ibuprofen every six hours'') may need to be verified against a passage reporting a weight-normalized dose (``$5$--$10$ mg/kg body weight''), requiring unit conversion and population assumptions. A claim about elderly patients (``dose reduction is required'') may only be derivable by chaining two passages: one linking age to hepatic decline, another linking hepatic decline to dose adjustment. And a claim phrased in general terms may need to be checked against evidence drawn from a specific subpopulation (e.g., adult males aged 18--65), requiring the verifier to judge whether the generalization is licensed. These misalignments are neither retrieval failures nor simple verdict errors. They demand domain knowledge, cross-passage reasoning, and calibrated handling of evidence strength --- capabilities that do not follow automatically from a stronger retriever or a larger verifier. \Cref{sec:qualitative analysis} examines these capabilities by dimension and shows that they account for a substantial share of errors that end-to-end metrics fail to reveal.

\section{Quantitative Analysis}
The data imbalance issue heavily influences precision in MedExpert. In contrast, Recall remains stable in the balanced MedExpert-gold in \Cref{tab:all-datasets-results} because most clinician-found factual errors can not be automatically detected. We use MedExpert-gold for detailed analysis in the main paper and report MedExpert-full results in \Cref{sec:MedExpert Full Results}. All calculation maps Refuted to Has Error and others to No Error.

A key question is whether the identified error patterns are artifacts of outdated models or fundamental limitations. We run ablation experiments on retrievers and verifiers to find that these errors persist, even using SOTA retrievers and SOTA verifiers, as shown in \Cref{tab:ablation_side_by_side} and discussed below.
\begin{table}[t]
\caption{Ablation results on the MedExpert-gold subset:
(a) Verifier ablation with Qwen3 retriever.
(b) Retriever ablation with GPT-5.4 verifier.}
\centering
\small
\begin{tabular}{lcc@{\qquad}ccc}
\toprule
\multicolumn{3}{c}{\textbf{(a) Verifier ablation}} &
\multicolumn{3}{c}{\textbf{(b) Retriever ablation}} \\
\cmidrule(r){1-3} \cmidrule(l){4-6}
\textbf{Verifier} & \textbf{Recall} & \textbf{F1} &
\textbf{Retriever} & \textbf{Recall} & \textbf{F1} \\
\midrule
Mistral-Small-24B-Instruct-2501 & \textbf{34.9} & \textbf{42.6} &
MedCPT & 19.8 & 29.3 \\
Gemma3-27B        & 8.1 & 14.1  &
RRF-2  & 24.4 & 34.4 \\
MedGemma-27B      & 16.3 & 25.2  &
RRF-4  & 18.6  & 28.8 \\
Qwen3.6-27B     & 22.1 & 32.8 &
Qwen3   & \textbf{25.6} & \textbf{35.2} \\
Mistral-Small-4-119B-2603 & 19.8 & 29.6 &
& & \\
GPT-5.4           & 25.6 & 35.2&
& & \\
\bottomrule
\end{tabular}
\label{tab:ablation_side_by_side}
\end{table}

\subsection{Better Retrievers Can't Solve the Challenge}
\paragraph{Existing fine-tuned medical retrievers are limited.}
\label{sec:find the best retriever}
\begin{table}[t]
\caption{Claim-level 4-label percentage distribution across retrievers with GPT-5.4 verifier on MedExpert-gold subset.}

\centering
\small
\begin{tabular}{ccccc}
\toprule
\textbf{Retriever} & \textbf{Supported\%} & \textbf{Refuted\%} & \textbf{NE\%} & \textbf{Ambiguous\%} \\
\midrule
MedCPT & 65.0 & 5.0 & 26.0 & 4.0 \\
RRF-2  & 71.5 & 5.6 & 19.2 & 3.7 \\
RRF-4  & 75.2 & 4.6 & 16.2 & 4.0 \\
Qwen3   & \textbf{78.3} & 7.1 & \textbf{8.0} & 6.6 \\
\bottomrule
\end{tabular}
\label{tab:retriever_label_distribution}
\end{table}

As shown in \Cref{tab:retriever_label_distribution}, Qwen3 has the largest number of functional evidence by the lowest number of No Evidence labels (8.04\%), turning most NE labels from other retrievers to more decisive Supported (80.14\%). From an end-to-end performance perspective, Qwen3 is the strongest retriever with the highest Recall and F1 in \Cref{tab:ablation_side_by_side}, while MedCPT/RRFs lead to lower Recall (i.e., their evidence can't help find clinician-identified factual errors) and lower F1 scores, showing weaker overall verification performance.

\subsection{Better Search Can't Solve the Challenge}
\label{sec:google search discussion}

As shown in \Cref{tab:corpus_comparison_google}, replacing the static MEDIC corpus with larger live Google General Search and Google Scholar Search, or the merged Google Search corpus does not improve Recall and F1. In \Cref{sec:Google Search Example}, we show that even authoritative sources (e.g., uclahealth.org) can contain unfactual content, and blindly expanding the corpus cannot solve this challenge.

\subsection{Better Verifiers Can't Solve the Challenge}
\label{sec:find the best verifier}

\paragraph{Verifier capability does not follow benchmark ranking.}
\label{sec: benchmark order flipped}

\begin{wrapfigure}[29]{r}{0.5\linewidth}
  \centering
  \vspace{-\intextsep}
  {\setlength{\belowcaptionskip}{\abovecaptionskip}\setlength{\abovecaptionskip}{0pt}%
  \captionof{table}{Corpus comparison with GPT-5.4 verifier and Qwen3 retriever on MedExpert-gold subset. Google Search denotes the merged corpus of Google General Search and Google Scholar Search.}
  \label{tab:corpus_comparison_google}}
  {\small
  \begin{tabular}{lcc}
  \toprule
  \textbf{Corpus} & \textbf{Recall} & \textbf{F1} \\
  \midrule
  MEDIC                 & \textbf{25.6} & \textbf{35.2} \\
  Google Scholar Search & 15.1  & 23.9 \\
  Google General Search & 17.4 & 26.1 \\
  Google Search         & 15.1  & 23.4 \\
  \bottomrule
  \end{tabular}\par}
  \vspace{6pt}
  \includegraphics[width=\linewidth]{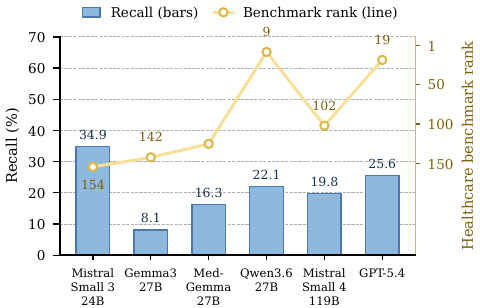}
  \stepcounter{footnote}%
  \caption{For each verifier model, bars (left axis) show recall on MedExpert-gold with the retriever fixed to Qwen3, and the line (right axis) shows its healthcare benchmark rank; models are ordered along the x-axis by size from small to large. Ranks are from the \href{https://llm-stats.com/leaderboards/best-ai-for-healthcare}{Best AI for Healthcare} leaderboard.\protect\footnotemark[\value{footnote}]}
  \label{fig:benchmark-recall-mismatch}
\end{wrapfigure}
\footnotetext[\value{footnote}]{MedGemma-27B is not tested by this leaderboard, but the \href{https://huggingface.co/google/medgemma-27b-it}{MedGemma model card} shows that its scores surpass Gemma3-27B. We therefore plot an estimated rank for it.}

As shown in \Cref{fig:benchmark-recall-mismatch}, the benchmark order usually scales by size, while we find that higher benchmark scores don't mean better end-to-end real-world performance (Recall/F1) in the open-ended medical verification task. Mistral Small 3 has the highest Recall with the smallest size, while GPT-5.4 and Qwen 3.6 have much lower Recall.
This mismatch suggests that standard healthcare benchmarks mainly reflect general medical QA or knowledge recall, but do not adequately measure a model's ability to use to verify open-ended clinical statements and detect subtle factual errors that occur during patient-facing communications. However, single metrics can not fully represent the intermediate capability of LLMs either, and we conducted large-scale, quantified qualitative analysis in \Cref{sec:verifier taxonomy main body} for deeper discussion.

\paragraph{Test-time scaling does not improve medical verification.}
\label{sec:adding reasoning test}
\suppressfloats[t]
\begin{table}[t]
\caption{GPT-5.4 reasoning effort ablation with Qwen3 retriever on MedExpert-gold subset.}

\centering
\small
\begin{tabular}{cccc}
\toprule
\textbf{Reasoning Effort} & \textbf{Recall} & \textbf{F1} & \textbf{Reasoning Tokens} \\
\midrule
Low    & \textbf{25.6} & 35.2 & 57,398 \\
Medium & 23.3 & 33.1 & 208,669 \\
High   & \textbf{25.6} & \textbf{37.3} & 466,052 \\
\bottomrule
\end{tabular}

\label{tab:gpt_reasoning_effort}

\end{table}

As shown in \Cref{tab:gpt_reasoning_effort}, increasing GPT-5.4's reasoning effort consumes substantially more reasoning tokens but does not improve recall. Instead, it wastes tokens on useless hedging, rather than leading to a correct verdict. This suggests that longer reasoning makes the verifier more hesitant, not more capable, in using evidence to determine whether a medical statement is factually wrong.

\paragraph{Medical fine-tuning yields limited improvements on Medical Verification.}
\label{sec:medical fine-tune marginal gain}

As shown in \Cref{tab:ablation_side_by_side}, MedGemma has roughly 2 times the Recall as Gemma, but still falls far behind the strongest verifier. This shows that fine-tuning cannot yield significant improvements beyond its base architecture and pretraining capabilities. F1 shows a similar overall performance gap.

\section{Qualitative Analysis}
\label{sec:qualitative analysis}

After 6 iterations, our codebooks are saturated, and adding new data can not yield any new patterns. We applied the final codebooks on the MedExpert-gold subset using Claude Opus 4.7 as a judge to label sub-codes for retrievers' 800 (claim, evidence) pairs and verifiers' 1686 (claim, reasoning) traces. We manually annotate 50 claim-evidence pairs and 60 claim-reasoning traces with human-LLM inter-annotator agreement of 0.86 and 0.8, respectively. 

\subsection{Verifier Behavior Taxonomy}
\label{sec:verifier taxonomy main body}

Our verifier behavior taxonomy decomposes the verification reasoning trace into 6 consecutive steps. The verifier starts by \textit{selecting} passages from the given evidence, and then \textit{interprets} its selected passages before conducting clinical \textit{inference}. After its inference and \textit{confidence} calibration, it \textit{concludes} a final label. All information in the reasoning text should be \textit{traceable} to the given evidence. Each step has a correct pattern and multiple error patterns in \Cref{tab:codebook_subcode_by_verifier}. 

We count each reasoning trace as correct only when all 6 steps are correct. Contrary to the end-to-end performance rank in \Cref{sec: benchmark order flipped}, GPT5.4 has the highest correctness rate (89\%), and Mistral Small3 has the lowest correctness rate (64\%), which shows that Mistral Small3's highest recall rate is mostly a lucky guess based on wrong reasoning traces. This again proves that relying on end-to-end aggregated metrics (e.g., Precision, Recall, F1) without multi-dimensional intermediate evaluation can be misleading in choosing the best models or systems.

Our quantified error rates (\%) show the dominant error patterns for each verifier, enabling cross-model comparison. Overall, Mistral Small 3 is the worst behavior model because it frequently ignores supporting or conflicting evidence passages (20.6\%) and fails to comprehend medical passages (16\%). Gemma3 misallocates attention to many irrelevant passages (6\%), while MedGemma mostly (1.1\%) anchors its attention to correct passages given the same evidence. In terms of confidence, all models except GPT-5.4 are usually overconfident based on the evidence strength (15.3\% to 9.6\%), while Mistral 3 and 4 models also have underconfident patterns (6\%, 5.3\%, respectively). GPT-5.4, as the closed-source SOTA model, still has unsolved intermediate error behaviors primarily at the inference and confidence steps, such as inferential overreach (6.4\%) and being overconfident based on the evidence strength (2.8\%). The open-source SOTA model, Qwen 3.6, has dominant errors in inferential overreach (12.8\%) and selective coverage of evidence (12.5\%).
The correctness rates (GPT5.4 $>$ Qwen3.6 $>$ Mistral Small4 $>$ MedGemma $>$ Gemma3 $>$ Mistral Small3) align with the current ranking of LLMs in terms of overall capability, with newer and larger models generally exhibiting stronger reasoning performance in the medical verification task.

\subsection{Retrieval Quality Taxonomy}
\label{sec:retriever taxonomy main body}

The final retrieval quality taxonomy in \Cref{tab:codebook_subcode_by_retriever} does not differ significantly from the seed codebook in \Cref{sec:error_extraction}, since it is the sixth version of the seed codebook that we synthesized comprehensive patterns from past iterations. 

We evaluate each passage's effectiveness for the medical fact verification task by 5 major dimensions: Entity Match, Scope Match, Content Trustworthiness, Evidence Polarity, and Chunking Issues\footnote{Temporal can be one dimension, but it is not added in the Table due to data limitations, acknowledge as one of our paper's Limitations}. Each dimension has one high-quality pattern and multiple low-quality patterns that degrade passage effectiveness in supporting a decision.

We count each passage as high quality only when it is high quality across all 5 dimensions. We calculate High Quality Rate, which is the portion of retrieved high-quality passages by each retriever, and find that aligns with conclusions in \Cref{sec:find the best retriever}: Qwen3 is the best retriever among all, and its passages' high quality rate (14.5\%) is 7 times the rate of the medically fine-tuned MedCPT retriever (2\%). However, most of Qwen3's retrieved evidence is still unsuitable to use. Current challenges for the SOTA Qwen3 retriever lie in finding the correct scope for medical conditions and populations, and locating direct signals for affirmation or contradiction.

\section{Promising Solutions}
\label{sec:proposed solutions}

Based on the deep analysis conducted on the open-ended retrieve-then-verify problem, we propose possible solutions as promising directions rather than a single prescriptive recipe: the design space is genuinely divergent, and in medical settings any solution must balance latency and token cost, against performance, rigor, and transparency. Promising directions include (a) intermediate-behavior evaluation based on our paper’s retrieval and verification taxonomies; (b) rubric-aware search agents whose retrieval and judgments are conditioned on rubrics derived from our error categories; (c) medical domain-knowledge injection and guideline-grounded rules as constraints or priors during verification; (d) medical expert-in-the-loop routing for low-confidence, high-stakes cases, which we recommend for real-world deployment. Our taxonomy provides the diagnostic foundation for future work to navigate performance and cost trade-offs in a targeted way.

\section{Conclusion}

Open-ended factuality evaluation requires a dedicated system with sufficient domain knowledge and capable reasoning skills. The most challenging problem is precisely locating all pieces of information scattered in different sources and aligning evidence with the claim, which will not automatically be solved as any single LLM scales up. For better retrieval quality, researchers should focus on customized task-oriented multi-dimensional evaluation in retrieving and re-ranking passages, rather than formula-based methods (e.g., lexical or cosine similarity in retrieving and RRF in reranking). To improve verification reliability, researchers should work on intermediate behavior evaluation to ensure verifiers are actually doing correctly, instead of competing on the aggregated metrics (e.g., P/R/F1).

\section*{Limitations}
\label{sec:limitation}
As of May 2026, MedExpert is the only publicly available open-ended medical factuality evaluation dataset with clinician-annotated True/False with clinician reasoning. MedRAG is the only publicly available corpus that includes a vast amount of authoritative medical passages, which was published in 2024. We build our pattern taxonomies and analysis mainly on this dataset and corpus. Since MedRAG does not provide metadata for each passage, we can not evaluate its passage publication date, which can be added in the Retriever Taxonomy (e.g., N/A for \textit{Temporal} Dimension). In the future, if other resources are publicly released, the Temporal information can be enriched.

Additionally, due to the open-endedness of the task, we do not have gold passage labels to calculate precision@k and recall@k for retriever metrics evaluation, nor do we have gold answers to calculate generator metrics for generative model evaluation, as proposed by RAGChecker \citep{ru2024ragchecker}. If more resources are released, researchers can further calculate these formula-based metrics to check if they align with the qualitative findings.

The automatic pattern tagging for percentage calculation is performed on the MedExpert-gold subset due to API budget limitations, and the percentage information can be more comprehensive by tagging more data.

\begin{ack}
This research was, in part, funded by the Advanced Research Projects Agency for Health (ARPA-H). The views and conclusions contained in this document are those of the authors and should not be interpreted as representing the official policies, either expressed or implied, of the United States Government.
\end{ack}
\bibliographystyle{plainnat}

\clearpage

\appendix
\section{Supplementary Details}
\label{sec:appendix}

\subsection{Full Retriever and Verifier Taxonomy Codebook}

\paragraph{Retriever Taxonomy.}
\Cref{tab:codebook_subcode_by_retriever} reports the high-quality rate per dimension rather than in aggregate, since a passage is counted as high quality on a dimension only when it meets the bar for that dimension in isolation.
For each error pattern, the retriever with the highest error rate is \textbf{bolded}; each retriever's dominant error is additionally \textit{italicized}.
In the metrics block, the \emph{High Quality Rate} is the fraction of passages that are simultaneously high quality across all dimensions; the best retriever per metric is \textbf{bolded}.

\paragraph{Verifier Taxonomy.} \Cref{tab:codebook_subcode_by_verifier} reports correctness per reasoning step rather than per claim, because end-to-end accuracy alone cannot certify trustworthiness in high-stakes settings.
Within a step, green and yellow sub-codes are mutually exclusive (a trace is either correct or it is not) but yellow sub-codes can co-occur, since a single trace may exhibit multiple distinct errors; per-step rows therefore sum to over 100\%.
For each error behavior, the verifier with the highest error rate is \textbf{bolded}; each verifier's dominant error is additionally \textit{italicized}.
In the metrics block, the best verifier per metric is \textbf{bolded}.

\subsection{Data Preprocessing and Alignment}
\label{sec:data preprocessing and alignment}
We applied dataset-specific preprocessing and verdict alignment procedures, including data cleaning, duplicate removal, data sampling, and label mapping, as described below.
\paragraph{MedExpert} We used MedExpert as the main open-ended evaluation dataset and sampled \textbf{MedExpert-gold} for manual case study evaluation.
To create MedExpert-gold, we selected all 225 sentences annotated as factually incorrect by domain experts in MedExpert. To ensure the ground truth quality, five clinicians (three in MH, two in PC) adjudicated the factuality of these sentences.

To assist the process, we provided the adjudicators the sentence-level claims, votes from six verifiers, and their reasoning. Of the 225 sentences, adjudicators retained 86 as containing factual errors: 40 agreed with the original annotation, 30 changed the severity level, 7 changed the description, 8 changed both the severity level and the description, and 1 changed the description and added a new omission error. The remaining 139 sentences were reclassified as factually correct: in 63 cases the factual error was removed, in 29 cases it was reclassified as an omission, in 8 cases the error was removed but a new omission error was introduced, and in 39 cases the original error was marked as a general comment only.

\textbf{Dataset Representativeness} 
\label{paragraph:representativeness}
Our work’s most findings are grounded on the MedExpert-gold subset that allows us to remain within a constrained budget without compromising on our finding’s representativeness. We developed this small scale dataset in an effort to mitigate the effects of low inter-annotator agreement reported by the original authors of MedExpert on our quantitative and qualitative analysis. Given that the original MedExpert dataset cost thousands of dollars in human annotation to make, we aim instead to create a sufficiently large but very high quality dataset for analysis. More specifically, 5 clinicians on our team manually created this subset of adjudicated annotations, to make sure our reported Precision/Recall/F1 is based on reliable annotations of the MedExpert-gold subset.

All original clinician annotated `Has Error` sentences are included in MedExpert-gold subset. We acknowledge that the subset’s Precision/F1 are not generalizable to the full set’s, but its Recall metric is since all of the `Has Error` examples remain. Since our paper’s main focus is to investigate whether retrieval-based verification systems can find the errors that clinicians find (i.e., Recall) and most of our qualitative and quantitative analyses focus on that metric, using this subset is a principled way to reduce cost. 

For human-machine verdict alignment, we used an any-error criterion: a sentence is treated as
has-error if at least one of its associated claims is labelled as false, and as
no-error otherwise. In the basic mapping setting, we mapped False to Refuted, while True was aligned with Supported, No Evidence, and Ambiguous.

\paragraph{SciFact} Since its official test set does not provide labels, we merged its training set and
validation set as our used dataset. This resulted in 1,109 labeled claims. We directly mapped the original claim-level verdicts to our four-label format: SUPPORT to Supported, CONTRADICT to Refuted, and Not Enough Info to No Evidence.

\paragraph{SciFact-Open} The original SciFact-Open set contained 279 claims. We identified three exact
duplicate claim strings:
(1) ``Hematopoietic progenitor cells are never susceptible to HIV-1 infection
ex vivo.'';
(2) ``Obesity decreases life expectancy.'';
and (3) ``Obesity prolongs life expectancy.''.
After removing duplicate claims, the final set contained 276 unique claims.

For claims associated with multiple evidence labels, we aggregated evidence-level verdicts into a single claim-level label. Claims with only SUPPORT evidence were mapped to Supported, claims with only CONTRADICT evidence were mapped to Refuted, and claims with only Not Enough Info evidence were mapped to No Evidence. If both SUPPORT and CONTRADICT evidence appeared for the same claim, the claim was mapped to Ambiguous.

\paragraph{HealthVer} The original HealthVer set contained 1,855 claims, where each claim may be
associated with multiple evidence sentences and each evidence sentence is
assigned an annotation label. During preprocessing, we
normalized claim strings by merging variants that differed only by leading or
trailing whitespace. No claims were removed as missing data; instead, the
corresponding rows were unioned under the same normalized claim. The affected
claims were:
(1) ``Children, like adults, who have COVID-19 but have no symptoms
(asymptomatic) can still spread the virus to others.'' 
(1 trailing-space variant row merged with 6 no-trailing-space rows);
(2) ``Drinking alcohol does not protect you against COVID-19 and can be
dangerous.'' (4 + 4 rows);
(3) ``Most people will experience a mild case with a 2-week recovery.''
(4 leading-space variant rows merged with 1 no-leading-space row);
and (4) ``Exposure to the sun or to temperatures higher than 77 F (25 C)
doesn't prevent the COVID-19 virus or cure COVID-19. You can get the COVID-19
virus in sunny, hot and humid weather.'' (3 + 1 rows).
After whitespace normalization, the dataset contained 1,851 normalized claims.

For verdict alignment, we aggregated the evidence-level labels of each normalized claim. Claims with Supports labels, with or without Neutral labels, were mapped to Supported; claims with Refutes labels, with or without Neutral labels, were mapped to Refuted; claims with only Neutral labels were mapped to No Evidence. Claims containing both Supports and Refutes labels were mapped to Ambiguous, regardless of whether Neutral labels were also present.

\paragraph{CovidFact} The original CovidFact set contained 4,086 claims. We identified five duplicate
claim strings:
(1) ``Sars-cov-2 viral load is associated with increased disease severity and
mortality'';
(2) ``Increased risk of noninfluenza respiratory virus infections associated
with receipt of inactivated influenza vaccine'';
(3) ``Fruitful neutralizing antibody pipeline brings hope to defeat
sars-cov-2'';
(4) ``Ivermectin exposure leads to up-regulation of detoxification genes in
vitro and in vivo in mice'';
and (5) ``Professional and home-made face masks reduce exposure to respiratory
infections among the general population''.
After removing duplicate claims, the final set contained 4,081 unique claims.

We directly mapped the original binary verdicts to our format, with SUPPORTED mapped to Supported and REFUTED mapped to Refuted.

\paragraph{Binary evaluation alignment}
Although our verification pipeline stores verdicts in a four-label format
(Supported, Refuted, No Evidence, and Ambiguous), we report precision, recall, and F1 in a binary True/False setup, to align with MedExpert clinician annotations. We have three binary label mapping strategies:
(1) only Refuted is treated as false, while Supported,
No Evidence, and Ambiguous are treated as true;
(2) Refuted and No Evidence are treated as false, while
Supported and Ambiguous are treated as true; and
(3) Refuted, No Evidence, and Ambiguous are treated as
false, while only Supported is treated as true.

\begin{table}[t]
\caption{Preprocessed sentence-level (MedExpert-related) and claim-level (Others) data statistics. NE means No Evidence (i.e., Neutral/Not Enough Info to decide a final Supported/Refuted verdict). Ambiguous means a claim has multiple pieces of supporting and refuting evidence.}
\centering
\begin{tabular}{lcccc}
\toprule
\textbf{Dataset}  & \textbf{\#Supported} & \textbf{\#Refuted} & \textbf{\#NE}  & \textbf{\#Ambiguous}\\
\midrule
MedExpert  & 9,950 & 224 & -& -\\
MedExpert-gold    & 139 & 86 & - & -\\
SciFact-Open  & 100 & 89 & 72 & 15\\
SciFact   & 456 & 237 & 416 & -\\
HealthVer   & 606 & 339 & 429 & 477\\
CovidFact   & 1,291 & 2,790 & - & -\\
\bottomrule
\end{tabular}%

\label{tab:data_statistics}
\end{table}

\definecolor{highQualityColor}{HTML}{9FE1CB}
\definecolor{lowQualityColor}{HTML}{F4E59F}
\definecolor{highQualityText}{HTML}{1F8A66}
\definecolor{lowQualityText}{HTML}{8A6A0F}

\definecolor{metricColor}{HTML}{E8EAED}
\definecolor{metricDark}{HTML}{4B5563}

\definecolor{cEntity}{HTML}{2563EB}
\definecolor{cScope}{HTML}{D97706}
\definecolor{cTrust}{HTML}{0891B2}
\definecolor{cPolarity}{HTML}{7C3AED}
\definecolor{cChunk}{HTML}{DC2626}

\definecolor{hlDark}{HTML}{1F2937}

\providecommand{\cellcolor}[1]{}

\providecommand{\errStrut}{\rule[-1.5pt]{0pt}{9pt}}
\providecommand{\errCC}[1]{\cellcolor{#1}\errStrut}
\providecommand{\errHF}[2]{\textbf{\textcolor{#1}{#2}}}                 
\providecommand{\errIT}[2]{\textit{\textcolor{#1}{#2}}}                 
\providecommand{\errBI}[2]{\textbf{\textit{\textcolor{#1}{#2}}}}        
\providecommand{\rotClass}[2]{\textcolor{#1}{\textbf{#2}}}

\gdef\errRowTint{}
\providecommand{\errNumCC}{\ifx\errRowTint\empty\else\cellcolor{\errRowTint}\fi}

\begin{table}[htbp]
\caption{\textbf{Evidence-quality breakdown across retrieval dimensions and end-to-end performance, using the best verifier, GPT-5.4.} \textcolor{highQualityText}{\textbf{Green}} sub-codes mark high-quality evidence patterns; \textcolor{lowQualityText}{\textbf{yellow}} marks imperfect evidence patterns. Each cell reports the \textbf{percentage} of the 200 annotated passages (per retriever) assigned that sub-code; percentages sum to 100 within each dimension. In the yellow rows, \textbf{bold} marks the highest rate across retrievers and \textit{italics} mark each retriever's dominant imperfect pattern. The bottom block reports end-to-end claim-verification performance (High Quality Rate, precision, recall, F1 with highest scores in bold \errHF{metricDark}{gray}) per retriever under GPT-5.4 verifier on the MedExpert-gold subset.}
\label{tab:codebook_subcode_by_retriever}

\centering
\small
\setlength{\tabcolsep}{2.5pt}
\renewcommand{\arraystretch}{1.15}
\renewcommand{\tabularxcolumn}[1]{m{#1}}  
\renewcommand{\errCC}[1]{\cellcolor{#1}\errStrut\gdef\errRowTint{#1!60}}
\begin{tabularx}{\textwidth}{
  >{\gdef\errRowTint{}\centering\arraybackslash}m{2.3cm}
  >{\raggedright\arraybackslash}m{2.5cm}
  >{\raggedright\arraybackslash}X
  >{\centering\arraybackslash\errNumCC}m{1.1cm}
  >{\centering\arraybackslash\errNumCC}m{1.1cm}
  >{\centering\arraybackslash\errNumCC}m{1.1cm}
  >{\centering\arraybackslash\errNumCC}m{1.1cm}
}
\toprule
\textbf{Dimension} & \textbf{Sub-code} & \textbf{Description}
  & \textbf{MedCPT} & \textbf{RRF2} & \textbf{RRF4} & \textbf{Qwen3} \\
\midrule

\multirow{3}{*}{\rotClass{cEntity}{\shortstack[c]{Entity\\Match}}}
 & Direct match
   & \errCC{highQualityColor} Exact entity + property match
   & 36.5 & 39.5 & 53 & 56.5 \\
\cmidrule(l){2-7}
 & Partial topical match
   & \errCC{lowQualityColor} Right property, related entity; Mechanism-only / pharmacological background
   & 28.5 & 31.5 & \errHF{hlDark}{32} & 29.5 \\
\cmidrule(l){2-7}
 & Irrelevant
   & \errCC{lowQualityColor} Same property, wrong domain; Wrong subdomain entirely
   & \errHF{hlDark}{35} & 29.0 & 15 & 14.0 \\
\midrule

\multirow{3}{*}{\rotClass{cScope}{Scope}}
 & Exact match
   & \errCC{highQualityColor} Scope precisely aligned with the claim
   & 7.5 & 9 & 13.5 & 27 \\
\cmidrule(l){2-7}
 & Partial match: Too broad/Too narrow
   & \errCC{lowQualityColor} Class-level for a specific claim; Subpopulation-specific
   & 57.5 & \errIT{hlDark}{62} & \errBI{hlDark}{71.5} & \errIT{hlDark}{56.0} \\
\cmidrule(l){2-7}
 & Not applicable
   & \errCC{lowQualityColor} Scope dimension does not apply (irrelevant passage)
   & \errHF{hlDark}{35} & 29.0 & 15 & 17 \\
\midrule

\multirow{2}{*}{\rotClass{cTrust}{Trustworthy}}
 & Reliable Source
   & \errCC{highQualityColor} Authoritative reference; Primary trial
   & 38 & 43.5 & 72 & 53.5 \\
\cmidrule(l){2-7}
 & Unreliable Source
   & \errCC{lowQualityColor} Non-peer-reviewed (opinions), or low-evidence-grade source
   & \errBI{hlDark}{62} & 56.5 & 28.0 & 46.5 \\
\midrule

\multirow{3}{*}{\rotClass{cPolarity}{\shortstack[c]{Evidence\\Polarity}}}
 & Direct affirmation or contradiction
   & \errCC{highQualityColor} Direct affirmation; Direct contradiction; Implicit entailment
   & 18.5 & 20.5 & 30 & 42 \\
\cmidrule(l){2-7}
 & Neutral or no-signal
   & \errCC{lowQualityColor} Topic mention without proposition
   & 46.5 & 50.5 & \errHF{hlDark}{55.0} & 41 \\
\cmidrule(l){2-7}
 & Not applicable
   & \errCC{lowQualityColor} Polarity dimension does not apply (irrelevant passage)
   & \errHF{hlDark}{35} & 29.0 & 15 & 17 \\
\midrule

\multirow{2}{*}{\rotClass{cChunk}{\shortstack[c]{Data ingestion \\ \& \\ Chunking issues}}}
 & Appropriate fragment/chunk
   & \errCC{highQualityColor} Well-formed passage with sufficient context
   & 96.5 & 100 & 91 & 95.5 \\
\cmidrule(l){2-7}
 & Incorrect fragmentation
   & \errCC{lowQualityColor} Truncated passage; Bad chunk boundaries
   & 3.5 & 0 & \errHF{hlDark}{9} & 4.5 \\
\midrule

\multirow{4}{*}{\textcolor{metricDark}{\textbf{Metrics}}}
 & \multirow{4}{*}{\makecell{Verifier:\\GPT5.4}}
   & \errCC{highQualityColor} High Quality Rate (all dimensions)
   & 2\% & 6\% & 11\% & \errHF{metricDark}{14.5\%} \\
\cmidrule(l){3-7}
 & & \errCC{metricColor} Precision
   & 56.7 & 58.3 & \errHF{metricDark}{64.0} & {56.4} \\
 & & \errCC{metricColor} Recall
   & 19.8 & 24.4 & 18.6 & \errHF{metricDark}{25.6} \\
 & & \errCC{metricColor} F1
   & 29.3 & 34.4 & 28.8 & \errHF{metricDark}{35.2} \\
\bottomrule
\end{tabularx}
\end{table}


\definecolor{highQualityColor}{HTML}{9FE1CB}
\definecolor{lowQualityColor}{HTML}{F4E59F}
\definecolor{highQualityText}{HTML}{1F8A66}
\definecolor{lowQualityText}{HTML}{8A6A0F}

\definecolor{metricColor}{HTML}{E8EAED}
\definecolor{metricDark}{HTML}{4B5563}

\definecolor{cInterp}{HTML}{6B5B3A}
\definecolor{cSelection}{HTML}{2D4A6A}
\definecolor{cInference}{HTML}{5B4A8A}
\definecolor{cLabel}{HTML}{2D5A3D}
\definecolor{cCalib}{HTML}{1A5878}
\definecolor{cGround}{HTML}{8A4A6A}

\definecolor{hlDark}{HTML}{1F2937}

\providecommand{\cellcolor}[1]{}

\providecommand{\errStrut}{\rule[-1.5pt]{0pt}{9pt}}
\providecommand{\errCC}[1]{\cellcolor{#1}\errStrut}
\providecommand{\errHF}[2]{\textbf{\textcolor{#1}{#2}}}
\providecommand{\errIT}[2]{\textit{\textcolor{#1}{#2}}}
\providecommand{\errBI}[2]{\textbf{\textit{\textcolor{#1}{#2}}}}
\providecommand{\rotClass}[2]{\textcolor{#1}{\textbf{#2}}}

\gdef\errRowTint{}
\providecommand{\errNumCC}{\ifx\errRowTint\empty\else\cellcolor{\errRowTint}\fi}

\onecolumn

\begingroup
\centering
\small
\setlength{\tabcolsep}{2.5pt}
\renewcommand{\arraystretch}{1.15}
\renewcommand{\errCC}[1]{\cellcolor{#1}\errStrut\gdef\errRowTint{#1!60}}

\begin{xltabular}{\textwidth}{
  >{\gdef\errRowTint{}\centering\arraybackslash}p{2cm}
  >{\raggedright\arraybackslash}p{2.5cm}
  >{\raggedright\arraybackslash}X
  >{\centering\arraybackslash\errNumCC}p{0.72cm}
  >{\centering\arraybackslash\errNumCC}p{0.72cm}
  >{\centering\arraybackslash\errNumCC}p{0.72cm}
  >{\centering\arraybackslash\errNumCC}p{0.72cm}
  >{\centering\arraybackslash\errNumCC}p{0.72cm}
  >{\centering\arraybackslash\errNumCC}p{0.72cm}
}

\caption{\textbf{Verifier-behavior breakdown across six sequential reasoning steps and end-to-end performance, using the best retriever, Qwen3.}
\textcolor{highQualityText}{\textbf{Green}} sub-codes mark correct verifier behaviors;
\textcolor{lowQualityText}{\textbf{yellow}} marks incorrect behaviors.
Each cell reports the \textbf{percentage} of the 281 evaluated claims (per verifier) on which the verifier flagged that sub-code; percentages can sum to more than 100 within each dimension because a single reasoning trace can contain multiple error sub-codes. In the yellow rows, \textbf{bold} marks the highest rate across verifiers and \textit{italics} mark each verifier's dominant error sub-code. 
The bottom block reports end-to-end claim-verification performance (Correctness Rate, precision, recall, F1 with highest scores in bold \errHF{metricDark}{gray}) per verifier under Qwen3 retriever on the MedExpert-gold subset.
\textcolor{red}{Red} numbers highlight the divergence between intermediate and end-to-end performance.}
\label{tab:codebook_subcode_by_verifier}\\

\toprule
\textbf{Step} & \textbf{Sub-code} & \textbf{Description}
  & \textbf{GPT} & \textbf{Qwen} & \textbf{MedG} & \textbf{Gem3} & \textbf{Mist3} & \textbf{Mist4} \\
\midrule
\endfirsthead

\multicolumn{9}{c}{\tablename~\thetable{} -- continued from previous page} \\
\toprule
\textbf{Step} & \textbf{Sub-code} & \textbf{Description}
  & \textbf{GPT} & \textbf{Qwen} & \textbf{MedG} & \textbf{Gem3} & \textbf{Mist3} & \textbf{Mist4} \\
\midrule
\endhead

\midrule
\multicolumn{9}{r}{Continued on next page} \\
\endfoot

\bottomrule
\endlastfoot

\multirow{3}{=}{\rotClass{cSelection}{\shortstack[c]{Evidence\\Selection}}}
 & Appropriate passage selection
   & \errCC{highQualityColor} Refers to all relevant passages provided in the Evidence in its Reasoning.
   & 96.4 & 86.8 & 85.8 & 78.3 & 77.6 & 89.7 \\
\cmidrule(l){2-9}
 & Selective coverage
   & \errCC{lowQualityColor} Partial Coverage; Ignores supporting or conflicting passage
   & 3.2 & 12.5 & 13.5 & 16.4 & \errBI{hlDark}{20.6} & 8.5 \\
\cmidrule(l){2-9}
 & Misallocated attention
   & \errCC{lowQualityColor} Anchors on off-topic passage; Over-weights low-quality source
   & 0.4 & 1.4 & 1.1 & \errHF{hlDark}{6.0} & 2.5 & 2.1 \\
\midrule

\multirow{2}{=}{\rotClass{cInterp}{\shortstack[c]{Evidence\\Interpretation}}}
 & Faithful read
   & \errCC{highQualityColor} Verbatim quote, correct use; Accurate paraphrase; Preserves qualifiers.
   & 98.6 & 95.0 & 89.3 & 85.8 & 84.0 & 90.4 \\
\cmidrule(l){2-9}
 & Comprehension failure
   & \errCC{lowQualityColor} Misreads clinical terminology
   & 1.4 & 5.0 & 10.7 & 14.2 & \errHF{hlDark}{16.0} & 9.6 \\
\midrule

\multirow{4}{=}{\rotClass{cInference}{\shortstack[c]{Inference\\Quality}}}
 & Sound inference
   & \errCC{highQualityColor} Multi-passage synthesis
   & 91.8 & 83.6 & 80.1 & 73.7 & 76.2 & 80.8 \\
\cmidrule(l){2-9}
 & Inferential overreach
   & \errCC{lowQualityColor} Non-contradictory-evidence-as-refuting
   & \errIT{hlDark}{6.4} & \errIT{hlDark}{12.8} & \errIT{hlDark}{14.2} & \errBI{hlDark}{22.1} & 18.1 & \errIT{hlDark}{15.7} \\
\cmidrule(l){2-9}
 & Conflict synthesis failure
   & \errCC{lowQualityColor} Fails to reconcile conflicting passages
   & 2.5 & 6.4 & 8.5 & \errHF{hlDark}{10.3} & 8.2 & 5.3 \\
\cmidrule(l){2-9}
 & Insufficient clinical reasoning
   & \errCC{lowQualityColor} Misses domain-obvious mechanism; Ignores clinical guidelines.
   & 1.8 & 1.8 & 2.5 & 1.8 & \errHF{hlDark}{3.2} & 2.8 \\
\midrule

\multirow{3}{=}{\rotClass{cCalib}{\shortstack[c]{Verbalized\\Model\\Confidence}}}
 & Appropriate confidence
   & \errCC{highQualityColor} Reasoning language matches the strength of the Evidence
   & 94.7 & 87.9 & 88.3 & 83.3 & 82.2 & 84.7 \\
\cmidrule(l){2-9}
 & Overconfident
   & \errCC{lowQualityColor} Reasoning language expresses greater certainty than the evidence warrants
   & 2.8 & 11.0 & 9.6 & \errHF{hlDark}{15.3} & 11.7 & 10.0 \\
\cmidrule(l){2-9}
 & Underconfident
   & \errCC{lowQualityColor} Reasoning language expresses greater uncertainty than the evidence warrants
   & 2.5 & 1.1 & 2.1 & 1.4 & \errHF{hlDark}{6.0} & 5.3 \\
\midrule

\multirow{2}{=}{\rotClass{cLabel}{\shortstack[c]{Label\\Consistency}}}
 & Consistent/Match
   & \errCC{highQualityColor} Final label matches the Model's self-described conclusion in its Reasoning.
   & 99.6 & 100 & 99.3 & 98.2 & 100 & 99.6 \\
\cmidrule(l){2-9}
 & Mismatch
   & \errCC{lowQualityColor} Reasoning contradicts assigned label
   & 0.4 & 0 & 0.7 & \errHF{hlDark}{1.8} & 0 & 0.4 \\
\pagebreak

\multirow{2}{=}{\rotClass{cGround}{Grounding}}
 & Grounded reasoning
   & \errCC{highQualityColor} All information traceable to evidence
   & 100 & 100 & 99.3 & 97.5 & 97.2 & 99.3 \\
\cmidrule(l){2-9}
 & Ungrounded reasoning
   & \errCC{lowQualityColor} Hallucinated information not in the given evidence
   & 0 & 0 & 0.7 & 2.5 & \errHF{hlDark}{2.8} & 0.7 \\
\midrule

\multirow{4}{=}{\textcolor{metricDark}{\textbf{Metrics}}}
 & \multirow{4}{=}{\shortstack{Retriever:\\Qwen3}}
   & \errCC{highQualityColor} Correctness Rate, all 6 steps
   & \errHF{metricDark}{0.89} & 0.79 & 0.71 & 0.67 & \textcolor{red}{0.64} & 0.75 \\
\cmidrule(l){3-9}
 & & \errCC{metricColor} Precision
   & \textcolor{red}{56.4} & \errHF{metricDark}{63.3} & 56.0 & 53.8 & {54.5} & 58.6\\
 & & \errCC{metricColor} Recall
   & \textcolor{red}{25.6} & 22.1 & 16.3 & 8.1 & \errHF{metricDark}{34.9} & 19.8\\
 & & \errCC{metricColor} F1
   & \textcolor{red}{35.2} & 32.8 & 25.2 & 14.1 & \errHF{metricDark}{42.6} & 29.6 \\

\end{xltabular}

\endgroup

\twocolumn
\onecolumn
\subsection{Closed-ended Datasets' Evidence Overlap}
\label{sec:closed-ended evidence overlap}

For the 4 closed-ended datasets with pre-defined evidence for each claim, we automatically checked how much of their evidence is included in the MEDIC corpus we use. 

We compute evidence overlap at the document-title level. Since MEDIC splits each paper into multiple passages while preserving the same title across passages, we first collapse MEDIC from passages to unique normalized paper titles. We apply the same title normalization to each benchmark evidence corpus, using headline for CovidFact, title for SciFact and SciFact-Open, and metadata.title for HealthVer. For each dataset, evidence overlap is computed as
$|\mathcal{T}_{eval} \cap \mathcal{T}_{MEDIC}| / |\mathcal{T}_{eval}|$,
where $\mathcal{T}_{eval}$ is the set of unique evidence titles in the benchmark corpus and $\mathcal{T}_{MEDIC}$ is the set of unique paper titles in MEDIC. We use the evaluation corpus as the denominator because the goal is to estimate MEDIC's retrievable evidence coverage for each benchmark, rather than to compare corpus sizes.

The evidence overlap statistics are reported in \Cref{tab:closed-ended-dataset_coverage}. The scientific datasets, SciFact and SciFact-Open, have the most evidence overlap (90.12\% and 81.46\%), thus making it still an unchallenging closed-ended task that can be solved as long as the evidence is retrieved. Since all SciFact and SciFact-Open claims are extracted from scientific abstracts, as long as the verifier can understand each claim's gold passage, it becomes an easy Natural Language Inference task that does not require cross-passage reasoning, evidence alignment, and complex domain knowledge. In contrast, CovidFact and HealthVer exhibit much lower evidence overlap with MEDIC, covering only 14.13\% and 41.72\% of their evidence titles, respectively. This makes them more challenging closed-ended benchmarks under our MEDIC-based retrieval setting. Unlike the two SciFact datasets, whose evidence is largely concentrated in scientific abstracts, CovidFact uses source articles linked from public web sources, while HealthVer draws its evidence from CORD-19 scientific articles and manually extracted abstract-level evidence statements (i.e., claims). Their native evidence spaces are therefore less covered by MEDIC. When the gold annotated evidence is missing from the scientific MEDIC corpus, verification must rely more on collectively reasoning from the 10 retrieved scientific passages and the verifier model's internal medical domain knowledge.

\begin{table}[htb]
\caption{Datasets' corpus statistics and coverage in MEDIC.}
\centering
\begin{tabular}{lrrrr}
\toprule
\textbf{Dataset} & \textbf{Records} & \textbf{Titles} & \textbf{In MEDIC} & \textbf{Coverage} \\
\midrule
covidfact & 2,806 & 2,789 & 394 & 14.13\% \\
scifact & 5,183 & 5,180 & 4,668 & 90.12\% \\
scifact-open & 12,236 & 12,233 & 9,965 & 81.46\% \\
healthver & 147,156 & 101,878 & 42,501 & 41.72\% \\
\bottomrule
\end{tabular}%

\label{tab:closed-ended-dataset_coverage}
\end{table}

\subsection{Retrieval and Verification Prompts}
\label{sec:retrieval and verification prompts}
\begin{table}[!t]
\caption{Retrieval and verification prompts used in our evaluation pipeline.}
\label{tab:retrieval-verification-prompts}
\centering
\begin{minipage}{1.0\linewidth}
\hrule
\vspace{0.5em}

\noindent \textbf{Qwen3 Retriever Prompt}

\begin{Verbatim}[
breaklines=true,
breaksymbolleft={},
breaksymbolright={},
baselinestretch=0.92
]
Instruct: Given a medical claim, retrieve relevant biomedical evidence passages that can verify or refute the claim.

Query: 
\end{Verbatim}
\vspace{0.7em}
\hrule
\vspace{0.7em}

\noindent \textbf{Verifier System Prompt}

\begin{Verbatim}[
breaklines=true,
breaksymbolleft={},
breaksymbolright={},
fontsize=\footnotesize,
baselinestretch=0.92
]
You are a medical expert verifying whether a medical claim is supported by retrieved evidence passages. You will be given a set of retrieved medical evidence passages and a claim to verify.

 Strict grounding rules: you MUST follow these:
- Base your judgment ONLY on the content of the provided evidence passages.
- If the retrieved passages do not contain information relevant to the claim, you MUST return "No Evidence".
- In the `reasoning` field, explicitly cite or quote the passage content that drove your decision.

Classify the claim using exactly one of the following labels:

- Supported: The retrieved evidence passages that are relevant to the claim collectively support it. The claim is factually correct according to the evidence.
- Refuted: The retrieved evidence passages contain information that directly contradicts the claim.
- No Evidence: The retrieved passages are topically irrelevant to the claim or do not provide sufficient information to support or refute it.
- Ambiguous: The evidence is mixed — some passages support the claim while others contradict it.

Your output must be a JSON object following this schema:
{
  "additionalProperties": false,
  "properties": {
    "reasoning": {"title": "Reasoning", "type": "string"},
    "label": {"enum": ["Supported", "Refuted", "No Evidence", "Ambiguous"], "title": "Label", "type": "string"}
  },
  "required": ["reasoning", "label"],
  "title": "VerificationResult", "type": "object"
}
\end{Verbatim}

\vspace{0.7em}
\noindent \textbf{Verifier User Prompt}

\begin{Verbatim}[
breaklines=true,
breaksymbolleft={},
breaksymbolright={},
baselinestretch=0.92
]
Verify the claim into **one** of the four labels, based on the given evidence.

You should output the reasoning first and then the label.

Evidence: {evidence}

Claim: {claim}
Output:
\end{Verbatim}

\vspace{0.3em}
\hrule
\end{minipage}
\end{table}

The full retrieval and verification prompts are shown in \Cref{tab:retrieval-verification-prompts}. We use zero-shot prompts for both retrieval and verification to test their plain performance. The Qwen3 retriever is instructed to retrieve biomedical passages that may support or refute each claim, aiming to maximize evidence relevance. The verifier then judges the claim with a four-label schema based on the retrieved evidence, allowing researchers to better analyze and reflect on models' ability to use external evidence for medical claim verification rather than only focusing on final labels and Precision/Recall/F1 metrics.

\subsection{Experiment Computation and Hyperparameters}
\label{sec:compute and hyperparameter}

\begin{table}[t]
\caption{Inference latency and computational cost of the retriever and verifier models.}
\label{tab:compute-cost}

\centering
\footnotesize

\setlength{\tabcolsep}{3.6pt}
\setlength{\extrarowheight}{0.8pt}
\renewcommand{\arraystretch}{1.16}

\resizebox{0.96\textwidth}{!}{%
\begin{tabular}{
@{}
>{\centering\arraybackslash}m{1.25cm}
>{\centering\arraybackslash}m{3.15cm}
>{\centering\arraybackslash}m{5.40cm}
>{\centering\arraybackslash}m{1.75cm}
>{\centering\arraybackslash}m{3.15cm}
@{}
}

\toprule

\textbf{Stage}
&
\textbf{Model}
&
\textbf{Experiment}
&
\textbf{Inference latency}
&
\textbf{Computational cost}
\\

\midrule


\multirow[c]{4}{1.25cm}[-0.55ex]{\centering\textbf{Retriever}}
&
MedCPT Query Encoder
&
\multirow[c]{4}{5.40cm}[-0.55ex]{\centering
Full MedExpert; \\27,751 claims; top-10 retrieval}
&
5h13m
&
\multirow[c]{3}{3.15cm}[-0.55ex]{\centering
1 CPU node with 300\,GB RAM}
\\

&
RRF-2
&
&
7h
&
\\

&
RRF-4
&
&
23h32m
&
\\

\cmidrule(lr){2-2}
\cmidrule(lr){4-5}

&
Qwen3-Embedding-8B
&
&
27h04m
&
1$\times$ NVIDIA A100 with 40\,GB VRAM
\\

\midrule


\multirow[c]{11}{1.25cm}[-0.55ex]{\centering\textbf{Verifier}}
&
Mistral-Small-24B-Instruct-2501
&
\multirow[c]{5}{5.40cm}[-0.55ex]{\centering
Full MedExpert; \\27,751 claims across four retrievers}
&
24h
&
\multirow[c]{4}{3.15cm}[-0.55ex]{\centering
1$\times$ NVIDIA A100 with 80\,GB VRAM}
\\

&
gemma-3-27b-it
&
&
22h56m
&
\\

&
medgemma-27b-text-it
&
&
55h
&
\\

&
Qwen3.6-27B
&
&
147h15m
&
\\

\cmidrule(lr){2-2}
\cmidrule(lr){4-5}

&
Mistral-Small-4-119B-2603
&
&
22h43m
&
2$\times$ NVIDIA H200 with 144\,GB VRAM each
\\


\cmidrule(lr){2-5}

&
\multirow[c]{3}{3.15cm}[-0.55ex]{\centering
GPT-5.4 (low)}
&
Four external datasets with Qwen3-Embedding-8B; 7,317 requests
&
6h20m
&
\$51.84
\\

\cmidrule(lr){3-5}

&
&
Full MedExpert with Qwen3-Embedding-8B; 27,751 requests
&
24h
&
\$149.82
\\

\cmidrule(lr){3-5}

&
&
MedExpert-gold across four retrievers; 2,708 requests
&
2h21m
&
\$15.19
\\


\cmidrule(lr){2-5}

&
GPT-5.4 (low)
&
\multirow[c]{3}{5.40cm}[-0.55ex]{\centering
MedExpert-gold reasoning-effort analysis; 677 requests}
&
35m
&
\$3.54
\\

&
GPT-5.4 (medium)
&
&
2h08m
&
\$4.33
\\

&
GPT-5.4 (high)
&
&
4h45m
&
\$5.66
\\

\bottomrule

\end{tabular}%
}

\end{table}

Following the findings from MedScore Appendix A.7 \citep{medscore}, we set the retrieval top-$k$ to 10 for all retrieval-based verification experiments, balancing evidence coverage and verification cost (e.g., running time, GPU costs and API credits). All retrieval experiments used MEDIC as the evidence corpus and returned 10 documents per claim. As shown in \Cref{tab:compute-cost}, MedCPT retrieval was run on a CPU node with 300GB RAM and took 5h13m for the full MedExpert set of 27,751 claims. RRF-2 and RRF-4 retrieval were run on the same CPU node and took 7h and 23h32m, respectively. Qwen3-Embedding-8B retrieval was run on one NVIDIA A100 Tensor Core 40GB GPU with an encoding batch size of 256 and took 27h04m.

All open-weight verifier models were hosted with vLLM \citep{kwon2023efficient}. For the full MedExpert set of 27,751 claims across four retrievers, google/gemma-3-27b-it was run on one NVIDIA A100 Tensor Core 80GB GPU and took 22h56m; google/medgemma-27b-text-it was run on one NVIDIA A100 Tensor Core 80GB GPU and took 55h; mistralai/Mistral-Small-24B-Instruct-2501 was run on one NVIDIA A100 Tensor Core 80GB GPU and took 24h; mistralai/Mistral-Small-4-119B-2603 was run on two NVIDIA H200 Tensor Core 144GB GPUs with tensor parallelism and took 22h43m; and Qwen/Qwen3.6-27B was run on one NVIDIA A100 Tensor Core 80GB GPU and took 147h15m. For all five open-weight verifier models, we used temperature 0.0, top-$p$ 1.0, a maximum output length of 4096 tokens, seed 42, and four-label JSON-schema structured output.

The GPT verifier experiments were run with GPT-5.4 through the OpenAI Responses API in Batch mode on a CPU node with 300GB RAM. For the main GPT verification runs, we used reasoning effort set to low, a maximum output length of 2048 tokens, and four-label JSON-schema structured output. The four closed-ended datasets with the Qwen
retriever contained 7,317 requests and cost \$51.84. The full MedExpert experiment with the Qwen retriever contained 27,751 requests and cost \$149.82. The 677-claim MedExpert subset with four retrievers contained 2,708 requests and cost \$15.19. The
677-claim Qwen-retriever reasoning-effort analysis used low, medium, and high settings, contained 2,031 requests, and cost \$13.53.

\subsection{Details of Taxonomy Construction and Validation Process}
\label{sec:taxonomy construction details}
\begin{table}[t]
\caption{Human--machine labelling agreement across the six verifier-behavior categories.}
\label{tab:verifier-label-agreement}
\centering
\footnotesize

\setlength{\tabcolsep}{8pt}
\renewcommand{\arraystretch}{1.20}

\begin{tabularx}{\textwidth}{
@{}
>{\centering\arraybackslash}X
>{\centering\arraybackslash}m{2.3cm}
>{\centering\arraybackslash}m{2.8cm}
>{\centering\arraybackslash}m{2.8cm}
@{}
}

\toprule
\textbf{Category}
&
\textbf{$N$}
&
\textbf{Raw agreement}
&
\textbf{$\kappa$ / $\alpha$}
\\

\midrule

Grounding
&
120
&
90.0\%
&
0.80
\\

Label Consistency
&
120
&
89.2\%
&
0.78
\\

Verbalized Model Confidence
&
180
&
81.1\%
&
0.56
\\

Evidence Interpretation
&
120
&
80.8\%
&
0.62
\\

Inference Quality
&
240
&
79.2\%
&
0.45
\\

Evidence Selection
&
180
&
66.7\%
&
0.26
\\

\midrule

\textbf{Pooled / macro}
&
\textbf{960}
&
\textbf{80.0\%}
&
\textbf{0.57}
\\

\bottomrule

\end{tabularx}

\end{table}

We adapt the open-sourced GitHub repository\footnote{\url{https://github.com/gaojie058/Act-onomy}} code and pipeline from \citet{gao2026interpretagentbehavior} for human-in-the-loop automatic pattern induction in the medical setting. Specifically, we adapt its pattern extraction tool into two versions, one for the retriever pattern induction 

and one for the verifier pattern induction.

For each version of codebook, we started from a manually synthesized seed codebook from the last iteration of the codebook, and then let the Claude Opus 4.7 model intake new data to see if there are emerging patterns  (e.g., leaf nodes) that are not included in the codebook, and add new patterns into the codebook. In each iteration, leaf nodes will be clustered if they belong to the same sub-code. If adding multiple rounds of new data does not bring in any new pattern, we consider this codebook to be saturated. If a leaf node is extremely rare (e.g., only occurs once) and does not belong to any sub-code in the final saturated version, we manually remove it from the final codebook. We iterated the full loop for 6 versions of the seed codebook to reach the final comprehensive pattern taxonomies for the retriever and verifier. The retriever taxonomy is saturated with 8080 passages (randomly sampled 202 claims$\times$ 4 retrievers$\times$ 10 passages), and the verifier taxonomy is saturated with 1212 reasoning traces (202 claims$\times$ 6 verifiers' reasoning traces).

For claim-evidence labelling, 5 human annotators each independently labeled 10 (claim, passage) pairs (N = 50) across five dimensions and 13 sub-codes (See \Cref{tab:codebook_subcode_by_retriever}). We compared Claude's annotations to the human labels using raw agreement, Cohen's $\kappa$, and Krippendorff's $\alpha$ (nominal). Pooled agreement was high at both the dimension level (raw = 91.2\%, $\kappa$ = 0.90, $\alpha$ = 0.90) and the subcode level (raw = 93.2\%, $\kappa$ = 0.86, $\alpha$ = 0.86); per-annotator dimension-level $\kappa$ ranged from 0.83 to 0.93.

For claim-reasoning labelling, we measure agreement between five annotators and the system across 60 reasoning traces on a 6-step / 16-sub-code taxonomy (See \Cref{tab:codebook_subcode_by_verifier}).  
We use $\alpha$ rather than Cohen's $\kappa$ because $\kappa$ is undefined whenever a rater's marginal is constant. The raw agreement between automated labels and human judgments is 80.0\%, pooled over 960 sub-code judgments (60 reasoning traces $\times$ 16 sub-codes). We also report chance-corrected agreement for a more rigorous assessment: pooled Cohen's $\kappa$ and Krippendorff's $\alpha$ are both 0.57, indicating moderate agreement. Raw agreement exceeds $\kappa$ because most sub-codes are sparse binary flags with skewed base rates; for example, \textit{Mismatch} is present in only approximately 2\% of traces, which mechanically affects the chance-agreement correction relative to raw agreement.

As shown in \Cref{tab:verifier-label-agreement}, the category-level breakdown indicates that this is not a uniform weakness but a specific and explainable pattern. Grounding and Label Consistency, the most safety-relevant categories, show substantial or near-substantial agreement ($\kappa = 0.80$ and $0.78$, respectively). The weakest category, Evidence Selection ($\kappa = 0.26$), reflects genuine subjectivity in judging which passages a model implicitly relied on from free-text reasoning, rather than a labelling error; this is a harder judgment than the binary/objective checks in Grounding or Label Consistency.

Per-annotator $\kappa$ ranges from 0.24 to 0.89 (annotator 1 = 0.89, annotator 2 = 0.71, annotator 3 = 0.57, annotator 4 = 0.38, and annotator 5 = 0.24), showing a wider spread than the claim-evidence task ($\kappa = 0.77$--$0.90$). We attribute this to reasoning-trace annotation requiring more interpretive judgment than evidence-passage annotation and note it as a limitation. Overall, the raw agreement is 80.0\% ($\kappa = \alpha = 0.57$, moderate) across six reasoning-quality categories, ranging from substantial agreement for Grounding and Label Consistency ($\kappa = 0.80$ and $0.78$) to fair agreement for Evidence Selection ($\kappa = 0.26$), reflecting the greater subjectivity of the latter.

\onecolumn

\subsection{Google Search Example}
\label{sec:Google Search Example}

\begingroup
\small
\setlength{\tabcolsep}{4pt}
\renewcommand{\arraystretch}{1.12}

\begin{longtable}{
>{\bfseries\raggedright\arraybackslash}p{0.15\textwidth}
>{\raggedright\arraybackslash}p{0.80\textwidth}
}
\caption{Representative example of corpus-dependent verification reasoning with GPT-5.4 verifier.}
\label{tab:google_search_example}
\\
\toprule
\multicolumn{2}{p{0.97\textwidth}}{
\textbf{Sentence:} Dizziness upon standing, \textcolor{red}{also known as orthostatic hypotension},
is relatively common during pregnancy and can be attributed to several physiological changes. \newline
\textbf{Claim:} Dizziness upon standing is also known as orthostatic hypotension. \newline
\textbf{Human annotation:} This is a specific diagnosis, not just dizziness with standing.
} \\
\midrule
\endfirsthead

\toprule
\textbf{Evidence} \\
\midrule
\endhead

\bottomrule
\endfoot

\bottomrule
\endlastfoot

\multicolumn{2}{l}{\textbf{MedRAG}} \\
\cmidrule(lr){1-2}

Evidence
&
\textbf{Title:} Orthostatic Hypotension -- Introduction.\par
\textbf{Text:} Orthostatic hypotension, also called postural hypotension, is a drop in blood pressure on standing to a certain extent. Orthostatic hypotension is defined as a sudden drop in blood pressure upon standing from a sitting or supine position. Clinically, this is diagnosed by a sustained reduction in systolic blood pressure of at least 20 mmHg or diastolic blood pressure of 10 mmHg within three minutes of standing after being supine for five minutes or at a 60-degree angle on the tilt table. Symptoms on presentation are commonly related to cerebral hypoperfusion, but patients can also be asymptomatic.
\\[0.3em]

&
\textbf{Title:} Orthostatic Hypotension -- Continuing Education Activity.\par
\textbf{Text:} Orthostatic hypotension is defined as a sudden drop in blood pressure upon standing from a sitting or supine position. Clinically, this is diagnosed by a sustained reduction in systolic blood pressure of at least 20 mmHg or diastolic blood pressure of 10 mmHg within three minutes of standing after being supine for five minutes or at a 60-degree angle on the tilt table. Symptoms on presentation are commonly related to cerebral hypoperfusion, but patients can also be asymptomatic.
\\[0.3em]

&
\textbf{Title:} Orthostasis (Archived) -- Introduction -- Clinical Relevance.\par
\textbf{Text:} When the autonomic compensatory mechanism is affected, there is a significant decrease of the central blood pressure, and symptoms of hypoperfusion can appear, e.g., dizziness, lightheadedness or syncope. This is called orthostatic hypotension.
\\

\cmidrule(lr){1-2}

Model Reason
&
\textbf{Reasoning:} The claim is contradicted by the evidence. The passages define orthostatic hypotension as a blood pressure drop on standing, not simply dizziness. For example: ``Orthostatic hypotension is defined as a sudden drop in blood pressure upon standing'' and is diagnosed by ``a sustained reduction in systolic blood pressure of at least 20 mmHg or diastolic blood pressure of 10 mmHg within three minutes of standing.'' Another passage says symptoms can occur from hypoperfusion, including ``dizziness, lightheadedness or syncope,'' meaning dizziness is a symptom of orthostatic hypotension, not another name for it. The evidence also notes that patients ``can also be asymptomatic,'' further showing dizziness is not equivalent to the condition.

\textbf{Label:} Refuted
\\

\specialrule{0.9pt}{3pt}{3pt}

\multicolumn{2}{l}{\textbf{Google General Search}} \\
\cmidrule(lr){1-2}

Evidence
&
\textbf{Search result 1.}\par
\textbf{Title:} Orthostatic hypotension - Cleveland Clinic.\par
\textbf{Link:} \url{https://my.clevelandclinic.org/health/diseases/9385-low-blood-pressure-orthostatic-hypotension}\par
\textbf{Content:} Orthostatic hypotension is a sudden drop in blood pressure when you stand from a seated or prone (lying down) position. You may feel dizzy or even faint.
\\[0.3em]

&
\textbf{Search result 2.}\par
\textbf{Title:} Dizziness - orthostatic hypotension | Better Health Channel.\par
\textbf{Link:} \url{https://www.betterhealth.vic.gov.au/health/conditionsandtreatments/dizziness-orthostatic-hypotension}\par
\textbf{Content:} Orthostatic hypotension is a form of low blood pressure. It happens when the blood vessels do not constrict (tighten) as you stand up.
\\[0.3em]

&
\textbf{Search result 3.}\par
\textbf{Title:} Orthostatic Hypotension - StatPearls - NCBI Bookshelf.\par
\textbf{Link:} \url{https://www.ncbi.nlm.nih.gov/books/NBK448192/}\par
\textbf{Content:} Orthostatic hypotension, also known as postural hypotension, is defined as a sudden drop in blood pressure that occurs upon standing from a ...
\\[0.3em]

&
\textbf{Search result 4.}\par
\textbf{Title:} Feel dizzy when you stand up? What it means and what to do about it.\par
\textbf{Link:} \url{https://www.uclahealth.org/news/article/feel-dizzy-when-you-stand-up-what-it-means-and-what-do-about}\par
\textbf{Content:} \textcolor{red}{Feeling dizzy when standing up (orthostatic hypotension)} is commonly caused by dehydration, medication or an underlying condition.
\\[0.3em]

&
\textbf{Search result 5.}\par
\textbf{Title:} Orthostatic hypotension (postural hypotension) - Diagnosis \& treatment.\par
\textbf{Link:} \url{https://www.mayoclinic.org/diseases-conditions/orthostatic-hypotension/diagnosis-treatment/drc-20352553}\par
\textbf{Content:} This form of low blood pressure might cause dizziness, lightheadedness or fainting when rising from sitting or lying down.
\\[0.3em]

&
\textbf{Search result 6.}\par
\textbf{Title:} Orthostatic Hypotension - Stanford Health Care.\par
\textbf{Link:} \url{https://stanfordhealthcare.org/medical-conditions/brain-and-nerves/orthostatic-hypotension.html}\par
\textbf{Content:} Orthostatic hypotension is a rapid and sudden decrease in blood pressure that occurs when a person changes position.
\\[0.3em]

&
\textbf{Search result 7.}\par
\textbf{Title:} Dizzy spells when you stand up: When should you worry?\par
\textbf{Link:} \url{https://www.health.harvard.edu/heart-health/dizzy-spells-when-you-stand-up-when-should-you-worry}\par
\textbf{Content:} Orthostatic hypotension is a drop in blood pressure when standing up. If it ever leads to loss of consciousness or a fall, ...
\\[0.3em]

&
\textbf{Search result 8.}\par
\textbf{Title:} Orthostatic hypotension - Wikipedia.\par
\textbf{Link:} \url{https://en.wikipedia.org/wiki/Orthostatic_hypotension}\par
\textbf{Content:} Orthostatic hypotension, also known as postural hypotension, is a medical condition wherein a person's blood pressure drops when they are standing up ...
\\[0.3em]

&
\textbf{Search result 9.}\par
\textbf{Title:} What is Orthostatic Hypotension?\par
\textbf{Link:} \url{https://www.pccoforegon.com/what-is-orthostatic-hypotension/}\par
\textbf{Content:} Orthostatic hypotension is also known his postural hypotension and commonly known as a ``head rush''. ... lightheaded, dizzy, or what is called a ``head rush''.
\\

\cmidrule(lr){1-2}

Model Reason
&
\textbf{Reasoning:} The evidence supports the claim that dizziness upon standing is also known as orthostatic hypotension. Multiple passages explicitly connect standing up, dizziness, and orthostatic hypotension: Cleveland Clinic says, ``Orthostatic hypotension is a sudden drop in blood pressure when you stand... You may feel dizzy,'' and UCLA Health states, ``Feeling dizzy when standing up (orthostatic hypotension)...'' Mayo Clinic likewise says this condition ``might cause dizziness, lightheadedness or fainting when rising from sitting or lying down.'' These passages show that dizziness upon standing is referred to as orthostatic hypotension.

\textbf{Label:} Supported
\\

\end{longtable}

\endgroup

\clearpage

\subsection{MedExpert Full Results}
\label{sec:MedExpert Full Results}
Verifier ablation results on the MedExpert Full set are reported in \Cref{tab:medexpert_full_verifier_ablation}.
\begin{table}[t]
\caption{Verifier ablation on MedExpert Full with Qwen3 retriever.}
\label{tab:medexpert_full_verifier_ablation}
\centering
\begin{tabular}{lcc}
\toprule
\textbf{Verifier} & \textbf{Recall} & \textbf{F1} \\
\midrule
Mistral-Small-24B-Instruct-2501 & \textbf{24.6} & 7.9 \\
Gemma3-27B        & 5.4 & 4.6 \\
MedGemma-27B      & 10.7 & 6.7 \\
Qwen3.6-27B       & 13.4 & 6.7 \\
Mistral-Small-4-119B-2603   & 12.9 & 6.6 \\
GPT-5.4           & 17.0 & \textbf{8.4} \\
\bottomrule
\end{tabular}
\end{table}

\subsection{Statistical Significance Analysis of Experimental Results}
\label{sec:1k-v.s.-gold subset}
\begin{table}[t]
\caption{Controlled ablation results with 95\% confidence intervals.}
\label{tab:controlled-ablation-ci}
\centering
\footnotesize

\setlength{\tabcolsep}{5pt}
\renewcommand{\arraystretch}{1.18}

\begin{tabularx}{\textwidth}{
@{}
>{\centering\arraybackslash}m{3.25cm}
>{\centering\arraybackslash}X
>{\centering\arraybackslash}m{3.0cm}
>{\centering\arraybackslash}m{3.0cm}
@{}
}

\toprule
\textbf{Experiment}
&
\textbf{Condition}
&
\textbf{Recall [95\% CI]}
&
\textbf{F1 [95\% CI]}
\\
\midrule

\multirow[c]{6}{=}{\centering
Verifier ablation\\
MedExpert-gold}
&
Mistral-Small-24B-Instruct-2501
&
34.9 [25.7, 45.4]
&
42.6 [33.1, 52.9]
\\

&
gemma-3-27b-it
&
8.1 [4.0, 15.9]
&
14.1 [6.2, 25.2]
\\

&
medgemma-27b-text-it
&
16.3 [10.0, 25.5]
&
25.2 [15.1, 36.7]
\\

&
Qwen3.6-27B
&
22.1 [14.6, 31.9]
&
32.8 [22.4, 44.4]
\\

&
Mistral-Small-4-119B-2603
&
19.8 [12.7, 29.4]
&
29.6 [20.0, 41.9]
\\

&
GPT-5.4
&
25.6 [17.5, 35.7]
&
35.2 [25.0, 46.3]
\\

\midrule

\multirow[c]{4}{=}{\centering
Retriever ablation\\
MedExpert-gold}
&
MedCPT Query Encoder
&
19.8 [12.7, 29.4]
&
29.3 [18.9, 40.7]
\\

&
RRF-2
&
24.4 [16.6, 34.5]
&
34.4 [24.6, 45.9]
\\

&
RRF-4
&
18.6 [11.8, 28.1]
&
28.8 [18.7, 40.7]
\\

&
Qwen3-Embedding-8B
&
25.6 [17.5, 35.7]
&
35.2 [25.0, 46.3]
\\

\midrule

\multirow[c]{4}{=}{\centering
Corpus comparison\\
MedExpert-gold}
&
MEDIC
&
25.6 [17.5, 35.7]
&
35.2 [25.0, 46.3]
\\

&
Google Scholar Search
&
15.1 [9.1, 24.2]
&
23.9 [14.0, 36.0]
\\

&
Google General Search
&
17.4 [10.9, 26.8]
&
26.1 [16.8, 38.3]
\\

&
Google Search
&
15.1 [9.1, 24.2]
&
23.4 [13.9, 35.3]
\\

\bottomrule
\end{tabularx}

\end{table}
\begin{table}[t]
\caption{Permutation Cochran's $Q$ tests of recall for controlled ablations. ``$>$'' denotes significantly higher recall in Holm-corrected pairwise comparisons.}
\label{tab:cochran-q-recall}
\centering
\footnotesize

\setlength{\tabcolsep}{5pt}
\renewcommand{\arraystretch}{1.20}

\resizebox{\textwidth}{!}{%
\begin{tabular}{
@{}
>{\centering\arraybackslash}m{3.35cm}
>{\centering\arraybackslash}m{2.00cm}
>{\centering\arraybackslash}m{2.00cm}
>{\centering\arraybackslash}m{8.25cm}
@{}
}

\toprule

\textbf{Experiment}
&
\textbf{$Q$ statistic}
&
\textbf{Omnibus $p$}
&
\textbf{Holm-significant pairwise contrasts}
\\

\midrule

\makecell[c]{Verifier ablation\\MedExpert-gold}
&
40.20
&
0.0005
&
Mistral-Small-24B-Instruct-2501 $>$ \{gemma-3-27b-it, medgemma-27b-text-it\};
Qwen3.6-27B $>$ gemma-3-27b-it;
GPT-5.4 $>$ gemma-3-27b-it
\\

\midrule

\makecell[c]{Retriever ablation\\MedExpert-gold}
&
3.32
&
0.3983
&
None
\\

\midrule

\makecell[c]{Corpus comparison\\MedExpert-gold}
&
8.32
&
0.0360
&
None
\\

\bottomrule

\end{tabular}%
}

\end{table}
\begin{table}[t]
\caption{Paired bootstrap analysis of F1 difference. $\Delta$F1 is defined as comparator minus reference; negative values favor the reference condition.}
\label{tab:paired-bootstrap-f1}
\centering
\footnotesize

\setlength{\tabcolsep}{4.5pt}
\renewcommand{\arraystretch}{1.18}

\begin{tabularx}{\textwidth}{
@{}
>{\centering\arraybackslash}m{3.15cm}
>{\centering\arraybackslash}m{3.75cm}
>{\centering\arraybackslash}X
>{\centering\arraybackslash}m{1.65cm}
>{\centering\arraybackslash}m{2.25cm}
@{}
}

\toprule
\textbf{Experiment}
&
\textbf{Reference}
&
\textbf{Comparator}
&
\textbf{$\Delta$F1 (pp)}
&
\textbf{Holm-adjusted $p$}
\\
\midrule

\multirow[c]{5}{=}{\centering
Verifier ablation\\
MedExpert-gold}
&
\multirow[c]{5}{=}{\centering
Mistral-Small-24B-Instruct-2501}
&
gemma-3-27b-it
&
$-28.4$
&
0.0010
\\

&
&
medgemma-27b-text-it
&
$-17.3$
&
0.0048
\\

&
&
Qwen3.6-27B
&
$-9.8$
&
0.1468
\\

&
&
Mistral-Small-4-119B-2603
&
$-13.0$
&
0.0660
\\

&
&
GPT-5.4
&
$-7.4$
&
0.1542
\\

\midrule

\multirow[c]{3}{=}{\centering
Retriever ablation\\
MedExpert-gold}
&
\multirow[c]{3}{=}{\centering
Qwen3-Embedding-8B}
&
MedCPT Query Encoder
&
$-5.9$
&
0.6791
\\

&
&
RRF-2
&
$-0.8$
&
0.8902
\\

&
&
RRF-4
&
$-6.4$
&
0.6791
\\

\midrule

\multirow[c]{3}{=}{\centering
Corpus comparison\\
MedExpert-gold}
&
\multirow[c]{3}{=}{\centering
MEDIC}
&
Google Scholar Search
&
$-11.3$
&
0.0832
\\

&
&
Google General Search
&
$-9.1$
&
0.1412
\\

&
&
Google Search
&
$-11.8$
&
0.0654
\\

\bottomrule
\end{tabularx}

\end{table}

To quantify the uncertainty associated with these experimental comparisons, Wilson confidence intervals for recall and stratified BCa bootstrap intervals for F1 are reported around each performance estimate. In addition, we use paired recall tests because each condition yields a detected-or-missed outcome on the same clinician-identified error-containing sentences. By contrast, F1 is analyzed separately because it depends jointly and nonlinearly on precision and recall rather than yielding a per-sentence binary outcome. Accordingly, paired F1 differences are assessed using a label-stratified bootstrap, which preserves the pairing across conditions while estimating uncertainty for this nonlinear dataset-level metric.

The results are shown in \Cref{tab:controlled-ablation-ci,tab:cochran-q-recall,tab:paired-bootstrap-f1}, no intervention yields a significant improvement. Once uncertainty and multiplicity are accounted for, none of the three interventions significantly improves recall or F1, and every significant effect favors the baseline configuration: the static MEDIC corpus significantly outperforms live web search on recall (Cochran's $Q$ $p = 0.0360$) and consistently on F1 ($\Delta$F1 up to $-11.8$ pp), and the baseline verifier has significantly higher F1 than the newer and medically-tuned Gemma alternatives ($p \leq 0.005$).

\subsection{Granularity Alignment Between System Outputs and Sentence-Level Labels}
\label{sec:granularity-alignment}
\begin{table}[t]
\caption{Alignment between clinician-provided error reasoning and system reasoning for Has Error sentences in MedExpert-gold.}
\label{tab:atomic-sentence-alignment}
\centering
\footnotesize

\setlength{\tabcolsep}{8pt}
\renewcommand{\arraystretch}{1.20}

\begin{tabularx}{\textwidth}{
@{}
>{\centering\arraybackslash}m{3.4cm}
>{\centering\arraybackslash}X
>{\centering\arraybackslash}X
>{\centering\arraybackslash}X
>{\centering\arraybackslash}m{3.6cm}
@{}
}

\toprule

\textbf{MedExpert-gold}
&
\textbf{System aligned}
&
\textbf{Partially aligned}
&
\textbf{Not aligned}
&
\shortstack{\textbf{System = all claims}\\\textbf{Supported}}
\\

\midrule

Has Error sentences (86)
&
38
&
0
&
10
&
38
\\

\bottomrule

\end{tabularx}

\end{table}

Atomic facts' system predictions are aggregated as the system prediction for the corresponding parent sentence. If the system predicts any claim as containing an error, the full sentence is also predicted as Has Error; only when all claims are predicted as Supported is the sentence predicted as No Error.

We also note that clinicians annotate sentences by highlighting spans and providing the reasoning associated with each detected error, which has the same granularity as the atomic facts over which the system makes predictions. To examine this alignment, for the Has Error sentences in MedExpert-gold, we manually reviewed all clinician-provided reasoning traces together with the corresponding claims' system reasoning to determine whether the system reasons aligned with the clinician reasons. We use the strictest label-mapping rule: Refuted, No Evidence, and Ambiguous are treated as False (Has Error), and only when all claims are Supported is the sentence predicted as True (No Error). As shown in \Cref{tab:atomic-sentence-alignment}, approximately 80\% of the system prediction reasons are aligned with the clinician reasons, while only 10 are not aligned.

\subsection{Examples for Taxonomy Sub-codes}
\label{sec:taxonomy detailed examples}
From the human-in-the-loop taxonomy induction methodology, we created saturated codebooks for multi-dimensional retrieval and verification evaluation. We provide detailed representative examples (i.e., leaf nodes) and quotes for each sub-error code below.


\definecolor{highQualityColor}{HTML}{9FE1CB}
\definecolor{lowQualityColor}{HTML}{F4E59F}
\definecolor{highQualityText}{HTML}{1F8A66}
\definecolor{lowQualityText}{HTML}{8A6A0F}

\definecolor{exnoteColor}{HTML}{C0392B}

\providecommand{\exnote}[1]{\textcolor{exnoteColor}{#1}}

\providecommand{\subcodeDesc}[2]{%
  \begingroup
  \setlength{\parskip}{1.5pt}%
  \setlength{\parindent}{0pt}%
  \textbf{#1}\par
  {#2}%
  \endgroup}

\newcolumntype{N}[1]{>{\raggedright\arraybackslash}p{#1}}
\newcolumntype{L}[1]{>{\raggedright\arraybackslash}p{#1}}

\providecommand{\twoex}[2]{%
  \setlength{\parskip}{3pt}%
  \setlength{\parindent}{0pt}%
  #1\par\vskip 3pt\hrule\par #2}

\newlength{\excolwd}

\providecommand{\ragex}[3]{%
  \begingroup
  \setlength{\parskip}{1pt}%
  \setlength{\parindent}{0pt}%
  \textbf{Claim:} \textit{``#1''}\par
  \textbf{Evidence:} \textit{``#2''}\space #3%
  \endgroup}

\subsubsection{Retriever-Behavior Examples}
\begingroup
\setlength{\tabcolsep}{5pt}
\renewcommand{\arraystretch}{1.20}
\raggedright

\setlength{\excolwd}{\dimexpr\linewidth - 4.8cm - 2\tabcolsep\relax}

\paragraph{Step 1: Entity Match.} Recall of entities from Evidence based on the claim.
\medskip
\resizebox{\linewidth}{!}{
\begin{tabular}{N{4.8cm} L{\excolwd} @{}}
\toprule
\textbf{Sub-code} & \textbf{Representative (claim, evidence) pairs} \\
\midrule
\cellcolor{highQualityColor}
\subcodeDesc{Direct match}{Evidence has complete overlap with the claim topic. The passage names the same entity (drug, condition, or procedure) as the claim.}
& \twoex
  {\ragex{Uterine rupture can pose serious risks to both mother and baby if not promptly addressed.}{Uterine rupture is a rare but a catastrophic event.}{\exnote{(passage names the exact condition from the claim)}}}
  {\ragex{An epidural can sometimes slow down labor progress.}{Does epidural anesthesia affect the course of labor and delivery?}{\exnote{(passage explicitly concerns the same intervention)}}} \\
\cmidrule(l){1-2}
\cellcolor{lowQualityColor}
\subcodeDesc{Partial topical match}{Some overlap such as a related entity, same mechanism, or pharmacological background. Evidence contains one entity from the claim but not all, or refers to the same drug class rather than the specific agent.}
& \twoex
  {\ragex{Reducing the dose of Prozac temporarily may be recommended.}{dose escalation of antidepressants}{\exnote{(same drug class but does not name Prozac/fluoxetine specifically)}}}
  {\ragex{The scan ensures the placenta is not too low in the uterus, a condition known as placenta accreta.}{Placenta previa with abnormal ultrasound appearance}{\exnote{(related placental condition, not placenta accreta)}}} \\
\cmidrule(l){1-2}
\cellcolor{lowQualityColor}
\subcodeDesc{Irrelevant}{Evidence does not contain any entities from the claim. The passage covers the wrong clinical domain or shares only superficial lexical overlap with the claim.}
& \twoex
  {\ragex{Closely monitoring the condition may be recommended.}{optimizing medical monitoring protocols}{\exnote{(surface overlap on ``monitoring'' but passage is off-topic)}}}
  {\ragex{Delivering via cesarean section may be recommended if the bleeding has been heavy or recurrent.}{history of postpartum hemorrhage (PPH) is a recognized risk factor for PPH in subsequent pregnancies}{\exnote{(shares obstetric domain but no entity overlap with the claim)}}} \\
\bottomrule
\end{tabular}}
 
\bigskip
 
\paragraph{Step 2: Scope.} Matching of claim's clinical modifiers such as Population, Intervention, Comparator, Outcome, Timeframe (PICOT). E.g., ``pregnant women over 30'', ``sertraline withdrawal symptoms after long-term use''.
\medskip
\resizebox{\linewidth}{!}{
\begin{tabular}{N{4.8cm} L{\excolwd} @{}}
\toprule
\textbf{Sub-code} & \textbf{Representative (claim, evidence) pairs} \\
\midrule
\cellcolor{highQualityColor}
\subcodeDesc{Exact match}{Evidence focus matches the claim perfectly with respect to PICOT.}
& \twoex
  {\ragex{Uterine rupture can pose serious risks to both mother and baby if not promptly addressed.}{evaluate risk factors, management, maternal and fetal outcomes of ruptured uterus}{\exnote{(passage scope aligns exactly with the claim's entity and clinical focus)}}}
  {\ragex{Dizziness upon standing is relatively common during pregnancy.}{102 clinical healthy pregnant women were tested with a modified orthostatic test over defined time periods during pregnancy}{\exnote{(population and outcome match the claim exactly)}}} \\
\cmidrule(l){1-2}
\cellcolor{lowQualityColor}
\subcodeDesc{Partial match: Too broad / Too narrow}{Evidence focus partially overlaps with the specified focus in the claim with respect to PICOT. E.g., a claim is about medication tolerance in adults but the evidence is about medication tolerance in children (Population).}
& \twoex
  {\ragex{Reducing the dose of Prozac temporarily may be recommended.}{Maximizing the dose of antidepressants is widely recommended in cases of non-response to medium-dose treatment.}{\exnote{(too broad: covers antidepressants generally rather than Prozac dose reduction)}}  }
  {\ragex{An epidural can sometimes slow down labor progress.}{Seventy healthy, nulliparous women, at more than 37 weeks' gestation with cervical dilatation from 2 to 4 cm}{\exnote{(too narrow: restricts population to nulliparous women, a subset not specified in the claim)}}} \\
\cmidrule(l){1-2}
\cellcolor{lowQualityColor}
\subcodeDesc{Not applicable}{This dimension is not applicable. Only to be used when the Entity Match is Irrelevant.}
& \twoex
  {\ragex{Closely monitoring the condition may be recommended.}{optimizing medical monitoring protocols}{\exnote{(Entity Match is Irrelevant; Scope dimension does not apply)}}}
  {\ragex{Delivering via cesarean section may be recommended if the bleeding has been heavy or recurrent.}{history of postpartum hemorrhage (PPH) is a recognized risk factor for PPH in subsequent pregnancies}{\exnote{(Entity Match is Irrelevant; Scope dimension does not apply)}}} \\
\bottomrule
\end{tabular}}
 
\bigskip
 
\paragraph{Step 3: Trustworthy.} Reliability of evidence for clinical decision-making. The hierarchy in medical decision-making runs from official guidelines and systematic reviews, to randomized controlled human clinical trials, to animal trials and case reports.
\medskip
\resizebox{\linewidth}{!}{
\begin{tabular}{N{4.8cm} L{\excolwd} @{}}
\toprule
\textbf{Sub-code} & \textbf{Representative (claim, evidence) pairs} \\
\midrule
\cellcolor{highQualityColor}
\subcodeDesc{Reliable Source}{Evidence considered strong, such as an authoritative reference from official guidelines, textbooks, or systematic reviews, or based on a randomized controlled trial with a sufficient number of subjects.}
& \twoex
  {\ragex{Reducing the dose of Prozac temporarily may be recommended.}{We performed a systematic literature search of Medline (1966--2003) and reviewed studies and publication references for available evidence.}{\exnote{(systematic review; high position in evidence hierarchy)}}}
  {\ragex{Uterine rupture can pose serious risks to both mother and baby if not promptly addressed.}{Population-based cohort study. Sweden. A total of 300,200 Swedish women delivering two single consecutive births between 1983 and 2001.}{\exnote{(large-scale population cohort with sufficient sample size)}}} \\
\cmidrule(l){1-2}
\cellcolor{lowQualityColor}
\subcodeDesc{Unreliable Source}{Evidence considered ``weak'' for clinical decision-making. E.g., a trial or study based on a small number of subjects, a case report, a study based on animals, or an opinion piece. Includes non-authoritative sources such as blogs and forums.}
& \twoex
  {\ragex{Uterine rupture can pose serious risks to both mother and baby if not promptly addressed.}{clinical data of 67 cases with uterine rupture in Woman's Hospital, School of Medicine, Zhejiang University were studied retrospectively}{\exnote{(small retrospective case series; limited generalizability)}}}
  {\ragex{Reducing the dose of Prozac temporarily may be recommended.}{Three drug dosing strategies can be employed to address dose-dependent drug adverse effects.}{\exnote{(opinion-style statement without cited trial data)}}} \\
\bottomrule
\end{tabular}}
 
\bigskip
 
\paragraph{Step 4: Evidence Polarity.} The ``stance'' of the evidence with respect to the claim.
\medskip
\resizebox{\linewidth}{!}{
\begin{tabular}{N{4.8cm} L{\excolwd} @{}}
\toprule
\textbf{Sub-code} & \textbf{Representative (claim, evidence) pairs} \\
\midrule
\cellcolor{highQualityColor}
\subcodeDesc{Direct affirmation or contradiction}{Evidence gives a direct affirmation or contradiction of the claim.}
& \twoex
  {\ragex{An epidural can sometimes slow down labor progress.}{The second stage of labor may be slightly prolonged by effective epidural analgesia}{\exnote{(directly affirms the claim)}}}
  {\ragex{Uterine rupture can pose serious risks to both mother and baby if not promptly addressed.}{Uterine rupture was associated with a substantially increased risk in neonatal mortality (adjusted OR 65.62; 95\% CI 32.60--132.08).}{\exnote{(directly supports the claim with quantitative evidence)}}} \\
\cmidrule(l){1-2}
\cellcolor{lowQualityColor}
\subcodeDesc{Neutral or no-signal}{Topic mention without proposition. The passage discusses the relevant topic but does not affirm or contradict the claim.}
& \twoex
  {\ragex{The scan ensures the placenta is not too low in the uterus, a condition known as placenta accreta.}{Accurate prenatal imaging is crucial for recognizing cases of the placenta accreta spectrum and for planning the necessary surgery.}{\exnote{(mentions the relevant condition but takes no stance on the claim)}}}
  {\ragex{An epidural can sometimes slow down labor progress.}{Is it possible to provide optimal analgesia throughout stage two labor without tending to increase the risk of instrumental delivery?}{\exnote{(poses a question about the topic without a propositional stance)}}} \\
\cmidrule(l){1-2}
\cellcolor{lowQualityColor}
\subcodeDesc{Not applicable}{This dimension is not applicable. Only to be used when the Entity Match is Irrelevant.}
& \twoex
  {\ragex{Delivering via cesarean section may be recommended if the bleeding has been heavy or recurrent.}{history of postpartum hemorrhage (PPH) is a recognized risk factor for PPH in subsequent pregnancies}{\exnote{(Entity Match is Irrelevant; polarity dimension does not apply)}}}
  {\ragex{Closely monitoring the condition may be recommended.}{optimizing medical monitoring protocols}{\exnote{(Entity Match is Irrelevant; polarity dimension does not apply)}}} \\
\bottomrule
\end{tabular}}
 
\bigskip
 
\paragraph{Step 5: Data Ingestion / Chunking Issues.} Issues arising when individual chunks have errors due to the segmentation of source documents.
\medskip
\resizebox{\linewidth}{!}{
\begin{tabular}{N{4.8cm} L{\excolwd} @{}}
\toprule
\textbf{Sub-code} & \textbf{Representative (claim, evidence) pairs} \\
\midrule
\cellcolor{highQualityColor}
\subcodeDesc{Appropriate fragment / chunk}{The chunk has enough context; no chunking or fragmentation error.}
& \twoex
  {\ragex{Closely monitoring the condition may be recommended.}{Monitoring is a form of surveillance consisting of repeated testing intended to detect a specified change in a patient indicating a change in his prognosis, need for treatment or need for a change in treatment.}{\exnote{(complete, self-contained passage with sufficient context)}}}
  {\ragex{Uterine rupture can pose serious risks to both mother and baby if not promptly addressed.}{Logistic regression was used to analyse potential risk factors for uterine rupture and risk of neonatal mortality associated with uterine rupture.}{\exnote{(passage retains enough context to evaluate the claim)}}} \\
\cmidrule(l){1-2}
\cellcolor{lowQualityColor}
\subcodeDesc{Incorrect fragmentation}{The passage is only a title, or context is lost when it was segmented from its source document.}
& \twoex
  {\ragex{Reducing the dose of Prozac temporarily may be recommended.}{TABLE 30--3 Antidepressant dose ranges. MAOIs, monoamine oxidase inhibitors; SNRIs, serotonin-norepinephrine reuptake inhibitors; SSRIs, selective serotonin reuptake inhibitors.}{\exnote{(table header only; no prose content retained from the source)}}}
  {\ragex{Delivering via cesarean section may be recommended if the bleeding has been heavy or recurrent.}{Cunningham FG, Nelson DB: Disseminated intravascular coagulation syndromes in obstetrics. Obstet Gynecol 126(5):999, 2015}{\exnote{(citation-only chunk; no substantive content available for evaluation)}}} \\
\bottomrule
\end{tabular}}

\bigskip

\endgroup

\subsubsection{Verifier-Behavior Examples}
\begingroup
\setlength{\tabcolsep}{5pt}
\renewcommand{\arraystretch}{1.20}
\raggedright

\setlength{\excolwd}{\dimexpr\linewidth - 4.8cm - 2\tabcolsep\relax}

\paragraph{Step 1: Evidence Selection.} Coverage of the Evidence cited in the model's Reasoning. (Note: selection is not an explicit step and is assumed based on the citations.)
\medskip
\resizebox{\linewidth}{!}{
\begin{tabular}{N{4.8cm} L{\excolwd} @{}}
\toprule
\textbf{Sub-code} & \textbf{Representative quotes from verifier reasoning} \\
\midrule
\cellcolor{highQualityColor}
\subcodeDesc{Appropriate passage selection}{Model refers to all relevant passages provided in the Evidence in its Reasoning. A ``coverage'' of the relevant passages.}
& \twoex
  {\textit{``The evidence passages consistently mention decaffeinated
 coffee\ldots\ The claim is supported by multiple converging passages.''}}
  {\textit{``The evidence passage titled `Human Trafficking -- Evaluation -- Reporting' explicitly states\ldots''}} \\
\cmidrule(l){1-2}
\cellcolor{lowQualityColor}
\subcodeDesc{Selective coverage}{Poor coverage of relevant passages from the Evidence. E.g., ignores supporting or conflicting passages, misses most relevant passage.}
& \twoex
  {\textit{``The claim is supported by multiple passages stating that
  monitoring/follow-up is recommended\ldots''} \exnote{(omits a passage
  that qualifies the cited evidence)}}
  {\textit{``None of the passages explicitly mention or evaluate drinking water.''} \exnote{(cherry-picks supporting passages while ignoring the on-topic gap)}} \\
\cmidrule(l){1-2}
\cellcolor{lowQualityColor}
\subcodeDesc{Misallocated attention}{Model's reasoning either does not cite a specific passage, or cites passages that are off-topic.}
& \twoex
  {\textit{``The passage titled `Protected Health Information -- Issues of Concern' states that PHI can be disclosed without consent in specific circumstances\ldots''} \exnote{(anchors on an off-topic passage)}}
  {\textit{``Several passages mention symptoms related to this drop in blood pressure, including dizziness.''} \exnote{(over-weights a tangential passage over the on-topic evidence)}} \\
\bottomrule
\end{tabular}}

\bigskip

\paragraph{Step 2: Evidence Interpretation.} Model understanding of a specific, cited passage.
\medskip
\resizebox{\linewidth}{!}{
\begin{tabular}{@{} N{4.8cm} L{\excolwd} @{}}
\toprule
\textbf{Sub-code} & \textbf{Representative quotes from verifier reasoning} \\
\midrule
\cellcolor{highQualityColor}
\subcodeDesc{Faithful read}{Model accurately interprets the Evidence through accurate paraphrasing, preservation of qualifiers, acknowledgment of gaps and competing evidence.}
& \twoex
  {\textit{``A significant increase in total fetal losses for early amniocentesis compared with midtrimester amniocentesis (7.6\% vs.\ 5.9\%).''} \exnote{(verbatim quote, qualifiers preserved)}}
  {\textit{``Patients may substitute decaffeinated for caffeinated coffee.''} \exnote{(accurate paraphrase)}} \\
\cmidrule(l){1-2}
\cellcolor{lowQualityColor}
\subcodeDesc{Comprehension failure}{Misunderstanding the Evidence due to a comprehension failure. E.g., misunderstands the Evidence hedging (``may'' to ``is''), flips the stance of the evidence, or over-generalizes the scope of conclusions present in the evidence.}
& \twoex
  {\textit{``Monitoring is recommended or appropriate for various medical conditions.''} \exnote{(clinical-construct conflation across unrelated passages)}}
  {\textit{``The claim incorrectly describes placenta previa as a condition where the placenta is *not* covering the cervix.''} \exnote{(misreads the claim text)}} \\
\bottomrule
\end{tabular}}

\bigskip

\paragraph{Step 3: Inference Quality.} Model interpretation of the selected Evidence cited in the model's Reasoning.
\medskip
\resizebox{\linewidth}{!}{
\begin{tabular}{@{} N{4.8cm} L{\excolwd} @{}}
\toprule
\textbf{Sub-code} & \textbf{Representative quotes from verifier reasoning} \\
\midrule
\cellcolor{highQualityColor}
\subcodeDesc{Sound inference}{Model accurately synthesizes information from multiple passages, including combining mixed evidence (supporting vs.\ refuting), recognizing insufficiency of evidence with respect to the claim, and linking relevant evidence for decision-making.}
& \twoex
  {\textit{``These passages collectively confirm that a temporary dose reduction of Prozac may be recommended.''} \exnote{(multi-passage  synthesis)}}
  {\textit{``The claim that switching to decaffeinated coffee is an option for reducing caffeine intake is factually correct.''} \exnote{(direct application)}} \\
\cmidrule(l){1-2}
\cellcolor{lowQualityColor}
\subcodeDesc{Inferential overreach}{Model makes a decision beyond what is provided in the evidence through generalizing sub-population findings to the population, or removal of hedging language (``may'' to ``is''). Includes unsupported conclusions not entailed by the evidence.}
& \twoex
  {\textit{``Multiple evidence passages support this claim.''} \exnote{(asserts proven from merely suggestive evidence)}}
  {\textit{``Assistance or information exchange can occur while maintaining some level of privacy.''} \exnote{(claim-reframing to fit the evidence)}} \\
\cmidrule(l){1-2}
\cellcolor{lowQualityColor}
\subcodeDesc{Conflict synthesis failure}{There are conflicting passages selected by the model but the final inference does not reconcile the conflicts.}
& \twoex
  {\textit{``One passage notes a higher rate (approximately 1\% to 3.9\%) specifically for women with prior cesarean sections\ldots''} \exnote{(cites one side of a conflict without reconciling the opposing passage)}}
  {\textit{``While the ideal scenario involves a discussion beforehand, the evidence suggests it doesn't always occur.''} \exnote{(notes the tension but does not resolve which interpretation is supported)}} \\
\cmidrule(l){1-2}
\cellcolor{lowQualityColor}
\subcodeDesc{Insufficient clinical reasoning}{Model fails to understand where the passage fits into the broader clinical understanding of the claim (e.g., ignores obvious standard-of-care guidelines because they are not explicitly mentioned), or fails on reasoning such as ``5\,mg'' being between ``5--20\,mg''.}
& \twoex
  {\textit{``All women capable of becoming pregnant should consume 400 micrograms of folic acid daily.''} \exnote{(treats a subgroup recommendation as a domain-general rule)}}
  {\textit{```Placenta previa' is a risk factor for extensive blood loss in both elective and emergency cesarean deliveries.''} \exnote{(misses the mechanistic distinction the claim turns on)}} \\
\bottomrule
\end{tabular}}

\bigskip

\paragraph{Step 4: Verbalized Model Confidence.} Hedging or confidence in the model's reasoning language for claim verification.
\medskip
\resizebox{\linewidth}{!}{
\begin{tabular}{@{} N{4.8cm} L{\excolwd} @{}}
\toprule
\textbf{Sub-code} & \textbf{Representative quotes from verifier reasoning} \\
\midrule
\cellcolor{highQualityColor}
\subcodeDesc{Appropriate confidence}{Reasoning language matches the strength (or lack thereof) of the Evidence for verifying the claim. Appropriately acknowledges uncertainty or certainty based on the evidence.}
& \twoex
  {\textit{``Switching to decaffeinated coffee is a viable option.''} \exnote{(hedge matches evidence strength)}}
  {\textit{``The claim is directly supported by the evidence.''} \exnote{(appropriate  certainty given decisive support)}} \\
\cmidrule(l){1-2}
\cellcolor{lowQualityColor}
\subcodeDesc{Overconfident}{The model's reasoning language expresses greater certainty than the evidence warrants. Asserts or strongly implies a claim is verified (or refuted) despite evidence that is partial, ambiguous, indirect, or insufficient.}
& \twoex
  {\textit{``Factually correct according to the evidence.''} \exnote{(hides available qualifiers)}}
  {\textit{``Directly supports the claim that it is possible to receive assistance (such as a consultation) without disclosing sensitive personal information.''} \exnote{(asserts proven from suggestive evidence)}} \\
\cmidrule(l){1-2}
\cellcolor{lowQualityColor}
\subcodeDesc{Underconfident}{The model's reasoning language expresses greater uncertainty than the evidence warrants. Hedges, qualifies, or withholds a verification judgment despite evidence that clearly and sufficiently supports (or refutes) the claim.}
& \twoex
  {\textit{``Not uniformly supported or refuted by the evidence.''} \exnote{(excessive hedging on otherwise decisive evidence)}}
  {\textit{``None of the passages address the scenario\ldots''} \exnote{(treats decisive evidence as No Evidence)}} \\
\bottomrule
\end{tabular}}

\bigskip

\paragraph{Step 5: Label Consistency.} Model's final label (claim verdict) based on its Reasoning.
\medskip
\resizebox{\linewidth}{!}{
\begin{tabular}{@{} N{4.8cm} L{\excolwd} @{}}
\toprule
\textbf{Sub-code} & \textbf{Representative quotes from verifier reasoning} \\
\midrule
\cellcolor{highQualityColor}
\subcodeDesc{Consistent/Match}{Final label matches the model's self-described conclusion in its Reasoning. E.g., ``There is no evidence supporting or refuting this claim'' $\rightarrow$ \texttt{No Evidence}.}
& \twoex
  {\textit{``The retrieved evidence explicitly supports the claim.''} \exnote{(final label aligns with reasoning conclusion)}}
  {\textit{``There is no mention of whether a breech presentation makes VBAC more or less suitable.''} \exnote{(reasoning $\rightarrow$ No Evidence alignment)}} \\
\cmidrule(l){1-2}
\cellcolor{lowQualityColor}
\subcodeDesc{Mismatch}{Model's assigned label contradicts its Reasoning. Includes labels conflicted by the reasoning (e.g., ``Passage 2 supports this claim'' $\rightarrow$ \texttt{No Evidence}) and inaccurate hedging (e.g., assigning \texttt{Ambiguous} to unidirectional evidence).}
& \twoex
  {\textit{``Therefore, the claim is refuted by the evidence\ldots''} \exnote{(assigned label is \texttt{Supported}; reasoning contradicts label)}}
  {\textit{``Bipolar disorder \textbf{can} be a debilitating long-term condition characterised by extreme mood swings.''} \exnote{(hedged reasoning paired with an unhedged label)}} \\
\bottomrule
\end{tabular}}

\bigskip

\paragraph{Step 6: Grounding.} Model relies on Evidence and not internal or fabricated information.
\medskip
\resizebox{\linewidth}{!}{
\begin{tabular}{@{} N{4.8cm} L{\excolwd} @{}}
\toprule
\textbf{Sub-code} & \textbf{Representative quotes from verifier reasoning} \\
\midrule
\cellcolor{highQualityColor}
\subcodeDesc{Grounded reasoning}{All cited information in the Reasoning is traceable to the Evidence.}
& \twoex
  {\textit{``Less than 17.7\,mg/dose.''} \exnote{(verbatim from the cited passage; all conclusions traceable to evidence)}}
  {\textit{``The passage titled `Creating supportive connections'\ldots''} \exnote{(explicit passage citation)}} \\
\cmidrule(l){1-2}
\cellcolor{lowQualityColor}
\subcodeDesc{Ungrounded reasoning}{Reasoning includes fabricated citations to non-existent passages, hallucinates quotes from passages, or relies on internal/parametric knowledge and ignores Evidence.}
& \twoex
  {\textit{``Water is a logical substitute.''} \exnote{(outside-knowledge injection; no passage grounds the conclusion)}}
  {\textit{``0--13.9\,mg per 16-oz serving.''} \exnote{(fabricated quote; figure does not appear in the retrieved passages)}} \\
\bottomrule
\end{tabular}}
\endgroup


\end{document}